\documentclass[twoside]{article}

\usepackage[accepted]{aistats2019}
\usepackage{microtype}
\usepackage{graphicx}
\usepackage{subfigure}
\usepackage{booktabs} % for professional tables
\usepackage{adjustbox} % for professional tables
\usepackage{multirow}
\usepackage{hyperref}
\hypersetup{
    colorlinks = true,
	citecolor=blue,
	linkcolor=blue
}

\usepackage{url}            % simple URL typesetting

\usepackage{amsmath,amssymb}
\usepackage{algorithmic,algorithm}
\usepackage{multirow}
\usepackage{comment}
\usepackage{enumerate}
\usepackage{wrapfig}
\usepackage{ascmac}
\usepackage{mdframed}
\usepackage{titling}
\usepackage{natbib}
\usepackage{framed}

\newcommand{\bs}{\boldsymbol}
\newcommand{\x}{\bs x}

\newcommand{\y}{\bs y}

\newcommand{\diff}{\mathrm{d}}

\newcommand{\fnn}{\bs f_{\text{NN}}}

\def\qed{\hfill $\Box$} 

\newtheorem{lem}{Lemma}[section]
\newtheorem{ex}{Example}[section]

\newtheorem{theo}{Theorem}[section]
\newtheorem{defi}{Definition}[section]
\newtheorem{prop}{Proposition}[section]

\allowdisplaybreaks[1]

\begin{document}



\twocolumn[

%%% 奥野タイトル案
\aistatstitle{Graph Embedding with Shifted Inner Product Similarity and Its Improved Approximation Capability}
%\aistatstitle{Graph Embedding using Shifted Inner Product Similarity \\ Goes Beyond Limitation}

%%% キム君タイトル案
%1. Beyond Inner Product Similarity
%2. Representation Learning with Shifted Inner Product Similarity
%3. Improved Representation Learning with Shifted Inner Product Similarity
%4. Graph Representation Learning with Shifted Inner Product Similarity
%5. Learning expressive distributed representations of graph

%%% キムくんはGeewook Kim (もしくはGee Wook Kimだが，本人はGeewook Kim希望)
\aistatsauthor{Akifumi Okuno$^{\dagger,\ddagger}$ \\ \url{okuno@sys.i.kyoto-u.ac.jp} \And Geewook Kim$^{\dagger,\ddagger}$ \\ \url{geewook@sys.i.kyoto-u.ac.jp} \And Hidetoshi Shimodaira$^{\dagger,\ddagger}$ \\ \url{shimo@i.kyoto-u.ac.jp}}

\aistatsaddress{
$^{\dagger}$Graduate School of Informatics, Kyoto University, \quad  $^{\ddagger}$RIKEN Center for Artificial Intelligence Project~(AIP)} 
]

\begin{abstract}
We propose \textit{shifted inner-product similarity~(SIPS)}, which is
 a novel yet very simple extension of the ordinary inner-product similarity~(IPS) for neural-network based graph embedding~(GE).
In contrast to IPS, that is limited to approximating positive-definite~(PD) similarities, 
SIPS goes beyond the limitation by introducing bias terms in IPS; we theoretically prove that SIPS is capable of approximating not only PD but also conditionally PD~(CPD) similarities with many examples such as cosine similarity, negative Poincar\'e distance and negative Wasserstein distance.
Since SIPS with sufficiently large neural networks learns a variety of similarities,  SIPS alleviates the need for configuring the similarity function of GE. 
Approximation error rate is also evaluated, and experiments on two real-world datasets demonstrate that graph embedding using SIPS indeed outperforms existing methods. 
\end{abstract}

\section{INTRODUCTION}
Graph embedding~(GE) of relational data, such as texts, images, and videos, etc., now plays an indispensable role in machine learning. 
To name but a few, words and contexts in a corpus constitute relational data, and their vector representations obtained by skip-gram model~\citep{mikolov2013distributed} and GloVe~\citep{pennington2014glove} are often used in natural language processing.
More classically, a similarity graph is constructed from data vectors, and nodes are embedded to a lower dimensional space where connected nodes are closer to each other~\citep{cai2018comprehensive}.

\begin{figure}[htbp]
\centering
	\includegraphics[width=0.48\textwidth]{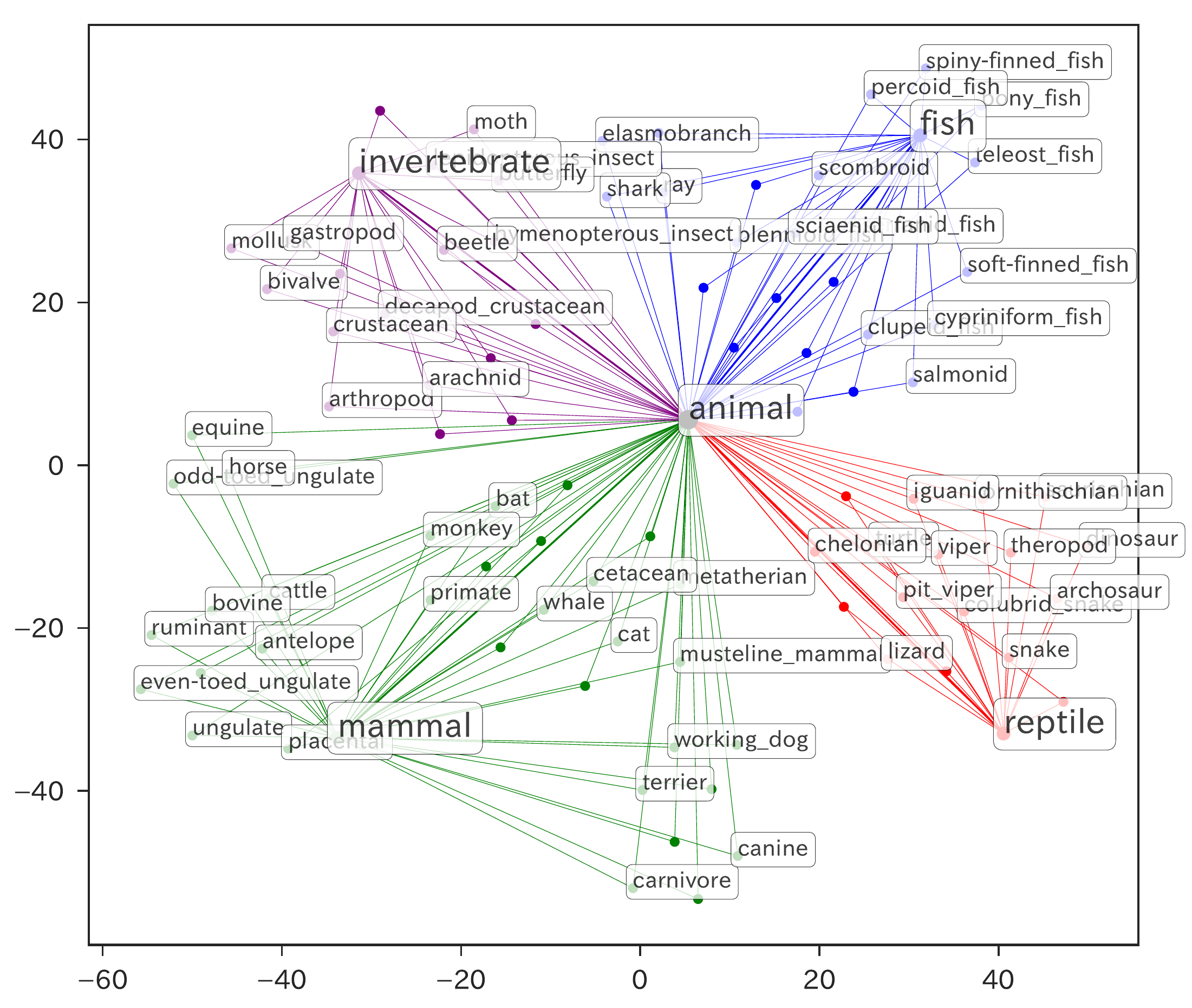}
	\caption{Visualization of word feature vectors for WordNet dataset computed by GE with our proposed SIPS model.
	See Supplement~\ref{sec:full_results_experiment} for details.	}
	\label{fig:wordnet}
\end{figure}

Embedding is often designed so that the inner product between two vector representations in Euclidean space expresses their similarity.
In addition to its interpretability, the inner product similarity has the following two desirable properties:
(1)~The vector representations are suitable for downstream tasks as feature vectors because machine learning methods are often based on inner products (e.g., kernel methods).
(2)~Simple vector arithmetic in the embedded space may represent similarity arithmetic such as
the ``linguistic regularities'' of word vectors \citep{mikolov2013linguistic}.
The latter property comes from the distributive law of inner product
$\langle \bs a + \bs b , \bs c \rangle = \langle \bs a, \bs c\rangle + \langle \bs b , \bs c \rangle$,
which decomposes the similarity of $\bs a + \bs b $ and $\bs c$ into the sum of the two similarities.
For seeking the word vector $\bs y' = \bs y_\text{queen}$, we maximize
$\langle \bs y_\text{king} - \bs y_\text{man} + \bs y_\text{woman}, \bs y' \rangle
= \langle \bs y_\text{king}, \bs y' \rangle - \langle \bs y_\text{man}, \bs y' \rangle + \langle \bs y_\text{woman}, \bs y' \rangle $ in Eq.~(3) of \citet{levy2014linguistic}.
Thus solving analogy questions with vector arithmetic is mathematically equivalent to seeking a word which is similar to king and woman but is different from man.

Although classical GE has been quite successful, it considers simply the graph structure, where data vectors (pre-obtained attributes such as color-histograms of images), if any, are used only through the similarity graph.
To fully utilize data vectors, neural networks~(NNs) are incorporated into GE so that data vectors are converted to new vector representations~\citep{kipf2016variational,zhang2017user,hamilton2017inductive,dai2018adversarial}, which reduces to the classical GE by taking 1-hot vectors as data vectors.
While these methods consider $1$-view setting, multi-view setting is considered in Probabilistic Multi-view Graph Embedding~\citep[PMvGE]{okuno2018probabilistic}, which generalizes existing multivariate analysis methods (e.g., PCA and CCA) and NN-extensions \citep[DCCA]{andrew2013deep} as well as graph embedding methods such as 
Locality Preserving Projections~\citep[LPP]{he2003locality,yan2007graph},
Cross-view Graph Embedding~\citep[CvGE]{huang2012cross}, and
Cross-Domain Matching Correlation Analysis~\citep[CDMCA]{shimodaira2016cross}.
In these methods, the inner product of two vector representations obtained via NNs represents the strength of association between the corresponding two data vectors.
The vector representations and the inner products are referred to as \emph{feature vectors} and \emph{Inner Product Similarities (IPS)}, respectively, in this paper.

IPS is considered to be highly expressive for representing the association between data vectors
due to the Universal Approximation Theorem~\citep[UAT]{funahashi1989approximate,cybenko1989approximation,
yarotsky2016error,pmlr-v70-telgarsky17a} for NN, which proves that NNs having many hidden units approximate arbitrary continuous functions within any given accuracy. 
However, since IPS considers the inner product of two vector-valued NNs, 
the UAT is \emph{not} directly applicable to the whole network with the constraints at the final layer.
Thus the approximation capability of IPS is yet to be clarified.

For that reason, \citet{okuno2018probabilistic} incorporates UAT into Mercer's theorem~\citep{minh2006mercer} and proves that IPS approximates any similarity based on Positive Definite~(PD) kernels arbitrary well. 
For example, IPS can learn cosine similarity, because it is a PD kernel.
This result shows not only the validity but also the fundamental limitation of IPS, meaning that the PD-ness of the kernels is required for IPS.

To overcome the limitation, similarities based on specific kernels other than the inner product have received considerable attention in recent years. 
One example is Poincar\'e embedding~\citep{nickel2017poincare} which is an NN-based GE using Poincar\'e distance for embedding vectors in hyperbolic space instead of Euclidean space.
Hyperbolic space is especially compatible with computing feature vectors of tree-structured relational data~\citep{sarkar2011low}. 
%Similarly, Gaussian embedding~\citep{vilnis2014word,bojchevski2018deep} is proposed to learn features based on Kullback-Leibler divergence. 
While these methods efficiently compute reasonable low-dimensional feature vectors by virtue of specific kernels, their theoretical differences from IPS is not well understood. 

%% theoretical insights on non-euclidean FLは言いすぎな気がする
In order to provide theoretical insights on these methods, in this paper, we will point out that some specific kernels are not PD by referring to existing studies. 
To deal with such non-PD kernels, we consider Conditionally PD~(CPD) kernels~\citep{berg1984harmonic,scholkopf2001kernel} which include PD kernels as special cases. 
We then propose a novel model named \emph{Shifted IPS~(SIPS)} that approximates similarities based on CPD kernels within any given accuracy. 
Interestingly, negative Poincar\'e distance is already proved to be CPD~\citep{faraut1974distances} and it is not PD. So, similarities based on this kernel can be approximated by SIPS but not by IPS.
Although we can think of a further generalization beyond CPD,
 this is only touched in Supplement~\ref{appendix:non_cpd} by defining \emph{inner product difference similarity~(IPDS)} model.

Our contribution is summarized as follows:
\begin{enumerate}[{(1)}]
\item We show that IPS cannot approximate a non-PD kernel; we propose SIPS to go beyond the limitation, and prove that SIPS can approximate any CPD similarities arbitrary well. 
\item We evaluate the error rate for SIPS to approximate CPD similarities, by incorporating neural networks such as multi-layer perceptron and deep neural networks.  
\item We conduct numerical experiments on two real-world datasets, to show that graph embedding using SIPS outperforms recent graph embedding methods.
\end{enumerate}

This paper is an extension of \citet{okuno2018representation} presented at ICML2018 workshop.

%\section{Background: Generative model for NN-based feature learning}
\section{BACKGROUND}
\label{sec:background}
We work on an undirected graph consisting of $n$ nodes $\{v_i\}_{i=1}^{n}$ and link weights $\{w_{ij}\}_{i,j=1}^{n} \subset \mathbb{R}_{\geq 0}$ satisfying $w_{ij}=w_{ji}$ and $w_{ii}=0$, where
$w_{ij}$ represents the strength of association between $v_i$ and $v_j$.
The data vector representing the attributes (or side-information) at $v_i$ is denoted as $\bs x_i \in \mathbb{R}^p$.
If we have no attributes, we use 1-hot vectors in $\mathbb{R}^n$ instead. 
We assume that the observed dataset consists of $\{w_{ij}\}_{i,j=1}^{n}$ and $\{\bs x_i\}_{i=1}^{n}$.

Let us consider a simple random graph model for the generative model of random variables $\{w_{ij}\}_{i,j=1}^{n}$ given data vectors $\{\bs x_i\}_{i=1}^{n}$.
The conditional distribution of $w_{ij}$ is specified by a \emph{similarity function} $h(\bs x_i, \bs x_j)$ of the two data vectors.
Typically, Bernoulli distribution $P(w_{ij}=1 | \bs x_i, \bs x_j) = \sigma(h(\bs x_i,\bs x_j))$
with sigmoid function $\sigma(x):=(1+\exp(-x))^{-1}$ for 0-1 variable $w_{ij}\in\{0,1\}$,
and Poisson distribution $w_{ij}\sim\text{Po}(\exp(h(\bs x_i,\bs x_j)))$ for non-negative integer variable $w_{ij}\in\{0,1,\ldots\}$
 are used to model the conditional probability.
These models are in fact specifying the conditional expectation $E(w_{ij}|\bs x_i, \bs x_j)$ by
$\sigma(h(\bs x_i,\bs x_j))$ and $\exp(h(\bs x_i,\bs x_j)) $, respectively, and they correspond to logistic regression and Poisson regression in the context of generalized linear models.

These two generative models are closely related.
Let $w_{ij} \sim \text{Po}(\lambda_{ij})$  with $\lambda_{ij}=\exp(h(\bs x_i,\bs x_j))$.
Then Supplement~\ref{appendix:equivalence_poisson_logistic} shows that
\begin{equation} \label{eq:poisson-sigmoid}
	P(w_{ij}=1 \mid \bs x_i, \bs x_j) = \sigma(h(\bs x_i,\bs x_j)) + O(\lambda_{ij}^3) 
\end{equation}
and $P(w_{ij}\ge2)=O(\lambda_{ij}^2)$, indicating that, for sufficiently small $\lambda_{ij}$,
the Poisson model is well approximated by the Bernoulli model.
Since these two models are not very different in this sense, we consider only the Poisson model in this paper.

We write the similarity function as
\begin{align}
h(\bs x_i,\bs x_j)
:=
g(\bs f(\bs x_i),\bs f(\bs x_j)), \label{eq:similarity}
\end{align}
where $\bs f:\mathbb{R}^p \to \mathbb{R}^K$ is a continuous function and $g:\mathbb{R}^{K \times K} \to \mathbb{R}$ is a symmetric continuous function.
For two data vectors $\bs x_i$ and $\bs x_j$, their feature vectors are defined as $\bs y_i = \bs f(\bs x_i)$ and $\bs y_j = \bs f(\bs x_j)$, thus the similarity function is also written as $g(\bs y_i, \bs y_j)$.
In particular, we consider a vector-valued neural network~(NN) $\bs y = \fnn(\bs x)$ for computing the feature vector, 
then $g(\fnn(\bs x_i),\fnn(\bs x_j))$ is especially called siamese network~\citep{bromley1994signature} in neural network literature. 
The original form of siamese network uses the cosine similarity for $g$, but we can specify other types of similarity function.
By specifying the inner product $g(\bs y, \bs y') = \langle \bs y, \bs y' \rangle$,
the similarity function (\ref{eq:similarity}) becomes
\begin{align}
h(\bs x_i,\bs x_j)
=
\langle \fnn(\bs x_i),\fnn(\bs x_j) \rangle.
\label{eq:ips}
\end{align}
We call (\ref{eq:ips}) as Inner Product Similarity~(IPS) model. 
IPS commonly appears in a broad range of methods, such as DeepWalk~\citep{perozzi2014deepwalk}, LINE~\citep{tang2015line}, node2vec~\citep{grover2016node2vec}, 
Variational Graph AutoEncoder~\citep{kipf2016variational}, and GraphSAGE~\citep{hamilton2017inductive}. 
Multi-view extensions \citep{okuno2018probabilistic} with views $d=1,\ldots, D$,
are easily obtained by preparing a neural network $\fnn^{(d)}$ for each view.

\section{PD SIMILARITIES}
\label{sec:pd_kernels}

In order to prove the approximation capability of IPS given in eq.~(\ref{eq:ips}), \citet{okuno2018probabilistic} incorporates the UAT for NN~\citep{funahashi1989approximate,cybenko1989approximation,yarotsky2016error,pmlr-v70-telgarsky17a} into Mercer's theorem~\citep{minh2006mercer}. 
In this section, we review their assertion that shows uniform convergence of IPS to any PD similarity. 
To show the result in Theorem~\ref{theo:universal_approximate}, we first define a kernel and its positive-definiteness.

\begin{defi}
\normalfont
\label{def:kernel}
For some set $\mathcal{Y}$,
    a symmetric continuous function $g:\mathcal{Y}^2 \to \mathbb{R}$ is called a \textit{kernel} on $\mathcal{Y}^2$. 
\end{defi}

\begin{defi}
\normalfont
\label{def:pd}
    A kernel $g$ on $\mathcal{Y}^2$ is said to be \textit{Positive Definite~(PD)} if satisfying $\sum_{i=1}^{n}\sum_{j=1}^{n} c_i c_j g(\bs y_i,\bs y_j) \geq 0$ for arbitrary $c_1,c_2,\ldots,c_n \in \mathbb{R},\bs y_1,\bs y_2,\ldots,\bs y_n \in \mathcal{Y}$. 
\end{defi}

For instance, cosine similarity 
$
g(\bs y,\bs y'):=\langle \frac{\bs y}{\|\bs y\|_2}, \frac{\bs y'}{\|\bs y'\|_2} \rangle
$ 
is a PD kernel on $(\mathbb{R}^p \setminus \{\bs 0\})^2$. 
Its PD-ness immediately follows from $\sum_{i=1}^{n}\sum_{j=c}^{n} c_i c_j g(\bs y_i,\bs y_j)=\|\sum_{i=1}^{n} c_i \frac{\bs y_i}{\|\bs y_i\|_2} \|_2^2 \geq 0$ for arbitrary $\{c_i\}_{i=1}^{n} \subset \mathbb{R}$ and $\{\bs y_i\}_{i=1}^{n} \subset \mathcal{Y}$. 
Also polynomial kernel, Gaussian kernel, and Laplacian kernel are PD~\citep{berg1984harmonic}.

\begin{defi}
\normalfont 
\label{def:similarity}
	A function $h(\bs x,\bs x'):=g(\bs f(\bs x),\bs f(\bs x'))$ with a continuous function $\bs f:\mathcal{X} \to \mathcal{Y}$ and a kernel $g:\mathcal{Y}^2 \to \mathbb{R}$ is called a \textit{similarity} on $\mathcal{X}^2$. 
\end{defi}

For a PD kernel $g$, the similarity $h$ is also a PD kernel on $\mathcal{X}^2$,
since  $\sum_{i=1}^{n}\sum_{j=1}^{n} c_i c_j h(\bs x_i,\bs x_j) = \sum_{i=1}^{n}\sum_{j=1}^{n} c_i c_j g(\bs f(\bs x_i),\bs f(\bs x_j)) \geq 0$.

Briefly speaking, a similarity $h$ is used for measuring how similar two data vectors are, while a kernel $g$ is used to compare feature vectors. 

The following theorem \citep{minh2006mercer} shows existence of a series expansion of any PD kernel,
which has been utilized in kernel methods in machine learning \citep{hofmann2008kernel}.

%decomposable は特別な意味があったりするので使いたくない

\begin{theo}[Mercer's theorem]
\normalfont
\label{theo:mercer}
For some compact set $\mathcal{Y} \subset \mathbb{R}^{K^*}$, $K^* \in \mathbb{N}$,
we consider a positive definite kernel $g_*:\mathcal{Y}^{2} \to \mathbb{R}$.
Then, there exist nonnegative eigenvalues $\{\lambda_k\}_{k=1}^{\infty}$, $\lambda_1\ge \lambda_2\ge\cdots$,
and continuous eigenfunctions $\{\phi_k\}_{k=1}^{\infty}$ such that
\begin{equation} \label{eq:mercer}
	g_*(\bs y_*,\bs y_*') = \sum_{k=1}^{\infty} \lambda_k \phi_k(\bs y_*)\phi_k(\bs y_*'),
\end{equation}
for all $\bs y_*, \bs y_*' \in \mathcal{Y}$, where
the series convergences absolutely for each $(\bs y_*,\bs y_*')$ and uniformly for $\mathcal{Y}$. 
\end{theo}

Note that the condition~(2) in \citet{minh2006mercer}, i.e.,
$\int_\mathcal{Y}\int_\mathcal{Y} g_*(\bs y_*, \bs y_*')\,d \bs y_*\, d\bs y_*'<\infty$, holds since $g_*$ is continuous and $\mathcal{Y}$ is compact. The theorem can be extended to closed set $\mathcal{Y}$, but we assume compactness for simplifying our argument.

It is obvious that IPS is always PD, because 
$	\sum_{i=1}^{n}\sum_{j=1}^{n} c_i c_j 
\big\langle \bs f_{\text{NN}}(\x_i), \bs f_{\text{NN}}(\x_j) \big\rangle
=\| \sum_{i=1}^n c_i \bs f_{\text{NN}}(\x_i) \|_2^2\geq 0$.
We would like to show the converse:~IPS approximates any PD similarities.
This is given by the Approximation Theorem (AT) for IPS below,
which is Theorem 5.1 $(D=1)$ of \citet{okuno2018probabilistic}.
The idea is to incorporate the UAT for NN into Mercer's theorem (Theorem~\ref{theo:mercer}).

\begin{theo}[AT for IPS] 
\label{theo:universal_approximate}
\normalfont
For $\mathcal{X} = [-M,M]^p$, $M>0$, and some compact set $\mathcal{Y} \subset \mathbb{R}^{K^*}$, $K^*\in\mathbb{N}$,
we consider a continuous function $\bs f_*:\mathcal{X} \to \mathcal{Y}$ and
a PD kernel $g^{\text{(PD)}}_*:\mathcal{Y}^2 \to \mathbb{R}$.
Let $\sigma(\cdot)$ be ReLU or an activation function which is non-constant, continuous, bounded, and monotonically-increasing.
Then, for arbitrary $\varepsilon>0$, 
by specifying sufficiently large $K \in \mathbb{N},m_f=m_f(K) \in \mathbb{N}$, 
there exist $\bs A \in \mathbb{R}^{K \times m_f},\bs B \in \mathbb{R}^{m_f \times p},\bs c \in \mathbb{R}^{m_f}$ such that
\begin{equation*}
\scalebox{1.0}{$
\bigg|
g^{\text{(PD)}}_*\left(\bs f_*(\x),\bs f_*(\x')\right)
-
\big\langle \bs f_{\text{NN}}(\x), \bs f_{\text{NN}}(\x') \big\rangle
\bigg|
<\varepsilon
$}
\end{equation*}
for all $(\bs x,\bs x') \in \mathcal{X}^2$, where $\bs f_{\text{NN}}(\x)
=
\bs A \bs \sigma(\bs B \x + \bs c)$ is a $1$-hidden layer neural network with $m_f$ hidden units and $K$ outputs, and $\bs\sigma(\x)$ is element-wise $\sigma(\cdot)$ function.
\end{theo}

See Supplement~A of \citet{okuno2018probabilistic} for the proof. 
It is
based on the series expansion
$g^{\text{(PD)}}_*(\bs y_*,\bs y_*') = \sum_{k=1}^{\infty} \lambda_k \phi_k(\bs y_*)\phi_k(\bs y_*')$
of Mercer's theorem (Theorem~\ref{theo:mercer}) for arbitrary PD kernel $g^{\text{(PD)}}_*$.
This expansion indicates with a vector-valued function 
$\tilde{\bs \phi}_K(\bs x):=(\lambda_1^{1/2}\phi_1(\bs f_*(\bs x)),\ldots,\lambda_K^{1/2}\phi_K(\bs f_*(\bs x)))$ that
$$
	\langle \tilde{\bs \phi}_K(\bs x), \tilde{\bs \phi}_K(\bs x') \rangle
	\to
	g^{\text{(PD)}}_*(\bs f_*(\bs x),\bs f_*(\bs x')), 
	\:
	K \to \infty,
$$
for all $\bs x,\bs x'$. 
Considering a vector-valued NN $\bs f_{\text{NN}}:\mathbb{R}^p \to \mathbb{R}^K$ that approximates $\tilde{\bs \phi}_K$, the IPS $\langle \bs f_{\text{NN}}(\bs x),\bs f_{\text{NN}}(\bs x') \rangle \approx \langle \tilde{\bs \phi}_K(\bs x),\tilde{\bs \phi}_K(\bs x') \rangle$ converges to $g^{\text{(PD)}}_*(\bs f_*(\bs x), \bs f_*(\bs x'))$ as $K \to \infty$, thus proving the assertion. 
In addition to the uniform convergence shown in Theorem~\ref{theo:universal_approximate},
the approximation error rate will be evaluated in Section~\ref{sec:error_bound_evaluation}.

Unlike Mercer's theorem which indicates only the existence of the feature map $\tilde{\bs \phi}_K$, 
Theorem~\ref{theo:universal_approximate} shows that a neural network $\bs f_{\text{NN}}:\mathbb{R}^p \to \mathbb{R}^K$ can be implemented so that the IPS $\langle \bs f_{\text{NN}}(\bs x),\bs f_{\text{NN}}(\bs x') \rangle$ eventually approximates the PD similarity $g^{\text{(PD)}}_*(\bs f_*(\bs x),\bs f_*(\bs x'))$  arbitrary well.

Note that Theorem~\ref{theo:universal_approximate} is AT for IPS which shows only the existence of NNs with required accuracy.
Although we do not go further in this paper,
consistency of the maximum likelihood estimation implemented as SGD is discussed in Section~5.2 and Supplement~B
of \citet{okuno2018probabilistic} for showing that IPS actually learns any PD similarities by increasing $n$.

\section{CPD SIMILARITIES}
\label{sec:cpd_kernels}

Theorem~\ref{theo:universal_approximate} shows that IPS approximates any PD similarities arbitrary well. 
However, similarities in general are not always PD.
To deal with non-PD similarities, we consider a class of similarities based on Conditionally PD~(CPD) kernels~\citep{berg1984harmonic,scholkopf2001kernel} which includes PD kernels as special cases. We then extend IPS to approximate CPD similarities.

Someone may wonder why only similarities based on inner product are considered in this paper.
In fact, it is obvious that a real-valued NN $f_\text{NN}(\bs x, \bs x')$ with sufficiently many hidden units approximates any similarity  $h(\bs x, \bs x')$ arbitrary well.
This is an immediate consequence of the UAT directly applied to $f_\text{NN}(\bs x, \bs x')$.
Therefore, considering the form $\langle \bs f_{\text{NN}}(\bs x),\bs f_{\text{NN}}(\bs x') \rangle $ or its extension just makes the problem harder.
Our motivation in this paper is that we would like to utilize the feature vector $\bs y = \bs f_{\text{NN}}(\bs x)$ with nice properties such as ``linguistic regularities'' which may follow from the constraint of the inner product.

The remaining of this section is organized as follows. 
In Section~\ref{subsec:limit_NNFL}, we point out the fundamental limitation of IPS to approximate a non-PD similarity.  
In Section~\ref{subsec:cpd_kernels}, we define CPD kernels with some examples. 
In Section~\ref{subsec:proposed_models}, we propose a novel Shifted IPS~(SIPS), by extending the IPS. 
In Section~\ref{subsec:interpretation}, we give interpretations of SIPS and its simpler variant C-SIPS.
In Section~\ref{subsec:rt_cpd}, we prove that SIPS approximates CPD similarities arbitrary well.

\subsection{Fundamental Limitation of IPS}
\label{subsec:limit_NNFL}

Let us consider the negative squared distance~(NSD) $g(\bs y,\bs y')=-\|\bs y-\bs y'\|_2^2$ and the identity map $\bs f(\bs x)=\bs x$. Then the similarity function
$$
h(\bs x,\bs x')
=
g(\bs f(\bs x),\bs f(\bs x'))
=
-\|\bs x-\bs x'\|_2^2 
$$
defined on $\mathbb{R}^{p} \times \mathbb{R}^p$ is not PD but CPD, which is defined later in Section~\ref{subsec:cpd_kernels}. 
Regarding the NSD similarity, Proposition~\ref{prop:dist_approximation_lower_bound} shows a strictly positive lower bound of approximation error for IPS.

\begin{prop}
\label{prop:dist_approximation_lower_bound}
\normalfont 
For all $M>0,p,K \in \mathbb{N}$, and a set of all $\mathbb{R}^K$-valued continuous functions $\mathfrak{S}(K)$, we have
\begin{align*}
&\inf_{\bs f \in \mathfrak{S}(K)}
\frac{1}{(2M)^{2p}}
\int_{[-M,M]^{p}}
\int_{[-M,M]^{p}} \\
&\hspace{2em}
\bigg|
	-\|\bs x-\bs x'\|_2^2
	-
	\langle
		\bs f(\bs x),\bs f(\bs x')
	\rangle
\bigg|
\diff \bs x
\diff \bs x'
\geq 
\frac{2pM^2}{3}.
\end{align*}
\end{prop}

The proof is given in Supplement~\ref{subsec:proof_lower_bound}.

Since $\mathfrak{S}(K)$ includes neural networks, Proposition~\ref{prop:dist_approximation_lower_bound} indicates that IPS does not approximate NSD similarity arbitrary well, even if NN has a huge amount of hidden units with sufficiently large output dimension.

\subsection{CPD Kernels and Similarities}
\label{subsec:cpd_kernels}
Here, we introduce similarities based on Conditionally PD~(CPD) kernels~\citep{berg1984harmonic,scholkopf2001kernel} in order to consider non-PD similarities which IPS does not approximate arbitrary well. 
We first define CPD kernels.

\begin{defi}
\normalfont 
	A kernel $g$ on $\mathcal{Y}^2$ is called \textit{Conditionally PD~(CPD)} if $\sum_{i=1}^{n}\sum_{j=1}^{n} c_i c_j g(\bs y_i,\bs y_j) \geq 0$ holds for arbitrary $c_1,c_2,\ldots,c_n \in \mathbb{R},\bs y_1,\bs y_2,\ldots,\bs y_n \in \mathcal{Y}$ with the constraint $\sum_{i=1}^{n} c_i=0$. 
\end{defi}

The difference between the definitions of CPD and PD kernels is whether it imposes the constraint $\sum_{i=1}^{n} c_i=0$ or not.
According to these definitions, CPD kernels include PD kernels as special cases.
For a CPD kernel $g$, the similarity $h$ is also a CPD kernel on $\mathcal{X}^2$.

A simple example of CPD kernel is $g(\bs y,\bs y')=-\|\bs y-\bs y'\|_2^{\alpha}$ for $0<\alpha \leq 2$ defined on $\mathbb{R}^K \times \mathbb{R}^K$. 
Other examples are $-(\sin(y-y'))^2$ and
$-\bs 1_{(0,\infty)}(y+y')$ on $\mathbb{R} \times \mathbb{R}$. 
CPD-ness is a well-established concept with interesting properties \citep{berg1984harmonic}:
For any function $u(\cdot)$, $g(\bs y,\bs y')=u(\bs y)+ u(\bs y')$ is CPD. Constants are CPD.
The sum of two CPD kernels is also CPD. 
For CPD kernels $g$ with $g(\bs y,\bs y')\le0$, CPD-ness holds for  $-(-g)^{\alpha} \: (\alpha \in (0,1])$ and $-\log(1-g)$.

\begin{ex}[Poincar\'e distance] 
\label{ex:poincare_embedding}
\normalfont 
For open unit ball $B^K:=\{ \bs y \in \mathbb{R}^K \mid \|\bs y \|_2 <1\}$,
we define a distance between $\bs y,\bs y' \in B^K$ as
\begin{equation}
\scalebox{0.85}{$\displaystyle 
d_{\text{Poincar\'e}}(\bs y,\bs y')
:=
\cosh^{-1}\left(
    1+2 \frac{\|\bs y-\bs y'\|_2^2}{(1-\|\bs y\|_2^2)(1-\|\bs y'\|_2^2)}
\right),$}
\label{eq:dist_poincare}
\end{equation}
where $\cosh^{-1}(z)=\log(z+\sqrt{z^2-1})$. 
Considering the generative model of Section~\ref{sec:background} with 1-hot data vectors, Poincar\'e embedding~\citep{nickel2017poincare} learns parameters $\bs y_i$, $i=1,\ldots,n$,
by fitting $\sigma(-d_{\text{Poincar\'e}}(\bs y_i, \bs y_j))$ to the observed $w_{ij} \in \{0,1\}$. 
Lorentz embedding~\citep{nickel2018learning} reformulate Poincar\'e embedding with a specific variable transformation, that enables more efficient computation. 
\end{ex}

Interestingly, negative Poincar\'e distance is proved to be CPD in \citet[Corollary 7.4]{faraut1974distances}. 

\begin{prop}
\label{prop:poincare_cpd}
\normalfont
$-d_{\text{Poincar\'e}}$ is CPD on $B^K \times B^K$. 
\end{prop}

%CPD-ness of $-d_{\text{Poincare}}$ is shown in  \citet{faraut1974distances} Corollary 7.4. 
$-d_{\text{Poincar\'e}}$ is strictly CPD in the sense that $-d_{\text{Poincar\'e}}$ is not PD.
A counter-example of PD-ness is, for example, $n=2,K=2,c_1=c_2=1,\bs y_1=(1/2,1/2),\bs y_2=(0,0) \in B^2$.

Another interesting example of CPD kernels is negative Wasserstein distance.

\begin{ex}[Wasserstein distance] 
\label{ex:wasserstein_embedding}
\normalfont 

Let $\bs Z$ be a metric space endowed with a metric $d_Z$, which we call as ``ground distance''. 
For $q\ge1$, let $\mathcal{Y}$ be the space of all measures $\mu$ on $\bs Z$  satisfying $\int_{\bs Z} d_Z(\bs z,\bs z_0)^q \diff \mu(\bs z) <\infty$ for some $\bs z_0 \in \bs Z$. 
The $q$-Wasserstein distance between $\bs y,\bs y' \in \mathcal{Y}$ is defined as
\begin{equation*}
\scalebox{0.9}{$ \displaystyle
d_{W}^{(q)}(\bs y,\bs y')
:=
\left(
	\inf_{\pi \in \Pi(\bs y,\bs y')}
	\iint_{\bs Z \times \bs Z}
	d_Z(\bs z,\bs z')^q
	\diff \pi(\bs z,\bs z')
\right)^{1/q}$}.
\end{equation*}
Here, $\Pi(\bs y,\bs y')$ is the set of joint probability measures on $\bs Z \times \bs Z$ having marginals $\bs y,\bs y'$. 
Wasserstein distance is used for a broad range of methods, such as Generative Adversarial Networks~\citep{arjovsky2017wasserstein} and AutoEncoder~\citep{tolstikhin2018wasserstein}. 
\end{ex}

Some cases of negative Wasserstein distance are proved to be CPD. 

\begin{prop}
\label{prop:wasserstein_cpd}
\normalfont
$-d_W^{(1)}$ is CPD on $\mathcal{Y}^2$ if $-d_Z$ is CPD on $\bs Z^2$. 
$-d_W^{(2)}$ is CPD on $\mathcal{Y}^2$ if $\bs Z$ is a subset of $\mathbb{R}$.
\end{prop}

$-d_W^{(1)}$ is known as the negative earth mover's distance, and its CPD-ness is discussed in \citet{gardner2017definiteness}. 
The CPD-ness of a special case of $-d_W^{(2)}$ is shown in \citet{kolouri2016sliced} Corollary~1. 
However, we note that negative Wasserstein distance, in general, is not necessarily CPD.
As Proposition~\ref{prop:wasserstein_cpd} states, $\bs Z$ is required to be a subset of $\mathbb{R}$ when considering $q>1$.

\subsection{Proposed Models}
\label{subsec:proposed_models}

For approximating CPD similarities, we propose a novel similarity model
\begin{align}
h(\bs x_i, \bs x_j) 
&=
\langle \bs f_{\text{NN}}(\bs x_i),\bs f_{\text{NN}}(\bs x_j) \rangle \nonumber \\
&\hspace{6em} + u_{\text{NN}}(\bs x_i) + u_{\text{NN}}(\bs x_j),
\label{eq:sips}
\end{align}
where $\bs f_{\text{NN}}:\mathbb{R}^p \to \mathbb{R}^K$ and $u_{\text{NN}}:\mathbb{R}^p \to \mathbb{R}$ are vector-valued and real-valued NNs, respectively. 
We call (\ref{eq:sips}) as Shifted IPS~(SIPS) model, 
because the IPS $\langle \bs f_{\text{NN}}(\bs x_i),\bs f_{\text{NN}}(\bs x_j) \rangle$ given in (\ref{eq:ips}) is
shifted by the offset $u_{\text{NN}}(\bs x_i) + u_{\text{NN}}(\bs x_j)$.
For illustrating how SIPS expresses CPD similarities, let us consider the NSD discussed in Section~\ref{subsec:limit_NNFL}:
\[
	-\|\bs x_i-\bs x_j\|_2^2=\langle \sqrt{2}\bs x_i,\sqrt{2}\bs x_j \rangle - \|\bs x_i\|_2^2 - \|\bs x_j\|_2^2
\]
is expressed by SIPS with $\bs f_{\text{NN}}(\bs x) = \sqrt{2}\bs x$ and $u_{\text{NN}}(\bs x) = - \|\bs x\|_2^2$.
Later, we show in Theorem~\ref{theo:universal_cpd} that SIPS approximates any CPD similarities arbitrary well.

We also consider a simplified version of SIPS.
By assuming $u_{\text{NN}}(\bs x)=-\gamma/2$ for all $\bs x$, SIPS reduces to 
\begin{align}
h(\bs x_i, \bs x_j) = 
\langle \bs f_{\text{NN}}(\bs x_i),\bs f_{\text{NN}}(\bs x_j) \rangle -\gamma, \label{eq:csips}
\end{align}
where $\gamma \in \mathbb{R}$ is a parameter to be estimated. 
We call (\ref{eq:csips}) as Constantly-Shifted IPS~(C-SIPS) model.

If we have no attributes, we use 1-hot vectors for $\bs x_i$ in $\mathbb{R}^n$ instead,
and $\bs f_{\text{NN}}(\bs x_i) = {\bs y}_i \in \mathbb{R}^K$, $u_{\text{NN}}(\bs x_i) = u_i \in \mathbb{R}$ are
model parameters. Then SIPS reduces to
the matrix decomposition model with biases
\begin{equation} \label{eq:matrix-factorization-with-bias}
h(\bs x_i, \bs x_j) = \langle {\bs y}_i, {\bs y}_j \rangle + u_i + u_j.
\end{equation}
This model is widely used for recommender systems \citep{koren2009matrix} and word embedding such as GloVe~\citep{pennington2014glove}, and SIPS is considered as its generalization.

\subsection{Interpretation of SIPS and C-SIPS}
\label{subsec:interpretation}
Here we illustrate the interpretation of the proposed models by returning back to the setting in Section~\ref{sec:background}. 
We consider a simple generative model of independent Poisson distribution with mean parameter $E(w_{ij}) = \exp(h(\bs x_i,\bs x_j))$.
Then SIPS gives a generative model
\begin{align}
\scalebox{0.9}{$\displaystyle 
w_{ij}
\overset{\text{indep.}}{\sim}
\text{Po}\Bigl(
	\beta(\bs x_i)
	\beta(\bs x_j)	
	\exp(\langle \bs f_{\text{NN}}(\bs x_i),\bs f_{\text{NN}}(\bs x_j) \rangle)\Bigr),  
	$}
	\label{eq:sips_weighted_model}
\end{align}
where $\beta(\bs x):=\exp(u_{\text{NN}}(\bs x))>0$. 
Since $\beta(\bs x)$ can be regarded as the ``importance weight'' of data vector $\bs x$, 
SIPS naturally incorporates the weight function $\beta(\bs x)$ to probabilistic models used in a broad range of existing methods.
Similarly, C-SIPS gives a generative model
\begin{align} 
w_{ij}
\overset{\text{indep.}}{\sim}
\text{Po}\Bigl(
	\alpha
	\exp(\langle \bs f_{\text{NN}}(\bs x_i),\bs f_{\text{NN}}(\bs x_j) \rangle) \Bigr),  \label{eq:csips_weighted_model}
\end{align}
where $\alpha:=\exp(-\gamma)>0$ regulates the sparseness of $\{w_{ij}\}$.
The generative model (\ref{eq:csips_weighted_model}) is already proposed as 1-view PMvGE~\citep{okuno2018probabilistic}.

It was shown in Supplement~C of \citet{okuno2018probabilistic} that PMvGE (based on C-SIPS) approximates CDMCA
when $w_{ij}$ is replaced by $\delta_{ij}$ in the constraint (8) therein,
and this result can be extended so that PMvGE with SIPS approximates the original CDMCA using $w_{ij}$ in the constraint.

\subsection{Approximation Theorems}
\label{subsec:rt_cpd}

It is obvious that SIPS is always CPD, because 
$	\sum_{i=1}^{n}\sum_{j=1}^{n} c_i c_j \big(
\langle \bs f_{\text{NN}}(\x_i), \bs f_{\text{NN}}(\x_j) \rangle
+ u_{\text{NN}}(\x_i) + u_{\text{NN}}(\x_j) \big)
=\| \sum_{i=1}^n c_i \bs f_{\text{NN}}(\x_i) \|_2^2
+ 2(\sum_{i=1} c_i) ( \sum_{j=1}c_j u_{\text{NN}}(\x_j) )
\geq 0$ for any $c_i$'s with $\sum_{i=1}^n c_i=0$.
We would like to show the converse:~SIPS approximates any CPD similarities,
and thus it overcomes the fundamental limitation of IPS.  
This is given in Theorem~\ref{theo:universal_cpd} below,
by extending Theorem~\ref{theo:universal_approximate} of IPS to SIPS.
Theorem~\ref{theo:universal_cpd2} also proves that C-SIPS given in eq.~(\ref{eq:csips})
 approximates CPD similarities in a weaker sense.

\begin{theo}[AT for SIPS]
\normalfont
\label{theo:universal_cpd}
For $\mathcal{X}=[-M,M]^{p}$, $M>0$, and some compact set $\mathcal{Y} \subset \mathbb{R}^{K^*}$, $K^*\in\mathbb{N}$,
we consider a continuous function $\bs f_*:\mathcal{X}\to \mathcal{Y}$ and
a CPD kernel $g^{\text{(CPD)}}_*:\mathcal{Y}^2 \to \mathbb{R}$.
Let $\sigma(\cdot)$ be ReLU or an activation function which is non-constant, continuous, bounded, and monotonically-increasing.
Then, for arbitrary $\varepsilon>0$, 
by specifying sufficiently large $K \in \mathbb{N},m_f=m_f(K)\in \mathbb{N},m_u \in \mathbb{N}$, 
there exist $\bs A \in \mathbb{R}^{K \times m_f},\bs B \in \mathbb{R}^{m_f \times p},\bs c \in \mathbb{R}^{m_f},
\bs e \in \mathbb{R}^{m_u}, \bs F \in \mathbb{R}^{m_u \times p}, \bs o \in \mathbb{R}^{m_u}$ such that
\begin{align*}
&\scalebox{0.9}{$
\bigg|
g^{\text{(CPD)}}_*\left(\bs f_*(\x), \bs f_*(\x')\right)$} \nonumber \\
&\hspace{3em}
\scalebox{0.9}{$
-
\left(\big\langle \bs f_{\text{NN}}(\x), \bs f_{\text{NN}}(\x') \big\rangle
+
u_{\text{NN}}(\x)
+
u_{\text{NN}}(\x')
\right)
\bigg|
<\varepsilon$}
\end{align*}
for all $(\bs x,\bs x') \in \mathcal{X}^2$, where $\bs f_{\text{NN}}(\x)
=
\bs A \bs \sigma(\bs B\x + \bs c) \in \mathbb{R}^{K}$ and
$u_{\text{NN}}(\x)
=
\langle \bs e, \bs \sigma(\bs F \x+\bs o) \rangle \in \mathbb{R}$ are one-hidden layer neural networks with $m_f$ and $m_u$ hidden units, respectively, 
and  $\bs\sigma(\x)$ is element-wise $\sigma(\cdot)$ function.
\end{theo}

The proof is in Supplement~\ref{sec:proof_universal_cpd}. 
It stands on Lemma 2.1 in \citet{berg1984harmonic}, which shows the equivalence of 
CPD-ness of $g^{\text{(CPD)}}_*(\bs y,\bs y')$ and PD-ness of
\begin{align}
g_0(\bs y,\bs y'):= &g_*^{\text{(CPD)}}(\bs y,\bs y') +g_*^{\text{(CPD)}}(\bs y_0,\bs y_0)
\nonumber \\
&\quad -g_*^{\text{(CPD)}}(\bs y,\bs y_0)  -g_*^{\text{(CPD)}}(\bs y', \bs y_0) \label{eq:g0cpd}
\end{align}
for any fixed $\bs y_0 \in \mathcal{Y}$. 
Using $g_0$ and $h_*(\bs x):=g_*^{\text{(CPD)}}(\bs f_*(\bs x),\bs y_0)-\frac{1}{2}g_*^{\text{(CPD)}}(\bs y_0,\bs y_0)$,
we write
\begin{align}
&g^{\text{(CPD)}}_*(\bs f_*(\bs x),\bs f_*(\bs x')) \nonumber \\
&\quad = g_0(\bs f_*(\bs x),\bs f_*(\bs x')) + h_*(\bs x) + h_*(\bs x').
\label{eq:gcpdg0}
\end{align}
AT for IPS shows that 
$\langle \bs f_{\text{NN}}(\bs x),\bs f_{\text{NN}}(\bs x') \rangle$ approximates
$g_0(\bs f_*(\bs x),\bs f_*(\bs x'))$ arbitrary well, and
UAT for NN shows that $u_{\text{NN}}(\bs x)$ approximates
$h_*(\bs x)$ arbitrary well, thus proving the theorem.

\begin{theo}[AT for C-SIPS]
\normalfont
\label{theo:universal_cpd2}
Symbols and assumptions are the same as those of Theorem~\ref{theo:universal_cpd}.
%記号はTheorem~\ref{theo:universal_cpd}と同様とする．
For arbitrary $\varepsilon>0$, 
by specifying sufficiently large $K \in \mathbb{N}$, $m_f=m_f(K) \in \mathbb{N}$, $r>0$, 
there exist $\bs A \in \mathbb{R}^{K \times m_f}$, $\bs B \in \mathbb{R}^{m_f \times p}$, $\bs c \in \mathbb{R}^{m_f}$, $\gamma=O(r^2)$ such that
\begin{align*}
&
\scalebox{0.9}{$
\bigg|
g^{\text{(CPD)}}_*\left(\bs f_*(\x), \bs f_*(\x')\right)$} \nonumber \\
&\hspace{3em}
\scalebox{0.9}{$-
\left(\big\langle \bs f_{\text{NN}}(\x), \bs f_{\text{NN}}(\x') \big\rangle
-\gamma\right)
\bigg|
<\varepsilon+O(r^{-2})$}
\end{align*}
for all $(\bs x,\bs x') \in \mathcal{X}^2$, where $\bs f_{\text{NN}}(\x)
=
\bs A \bs \sigma(\bs B\x + \bs c) \in \mathbb{R}^{K}$ is a one-hidden layer neural network with $m_f$ hidden units.
\end{theo}

The proof is in Supplement~\ref{sec:proof_universal_cpd2}. 

There is an additional error term of $O(r^{-2})$ in Theorem~\ref{theo:universal_cpd2}.
A large $r$ will reduce the error, but then 
large $\gamma=O(r^2)$ value may lead to unstable computation for finding an optimal NN.
Conversely, a small $r$ increases the upper bound of the approximation error . 
Thus, if available, we prefer SIPS in terms of both computational stability and small approximation error.

\section{APPROXIMATION ERROR RATE}
\label{sec:error_bound_evaluation}
Thus far, we showed universal approximation capabilities of IPS and SIPS in Theorems~\ref{theo:universal_approximate} and \ref{theo:universal_cpd}.
In this section, we evaluate error rates for these approximation theorems, by assuming some additional conditions.
They are used for employing the theorems for eigenvalue decay rate of PD kernels~\citep[Theorem~4]{cobos1990eigenvalues} and approximation error rate for NNs~\citep{yarotsky2018optimal}. 

\textbf{Conditions on the similarity function:}
We consider the following conditions on 
the function $\bs f_*$ and the kernel $g_*$ for the underlying true similarity $g_*(\bs f(\x), \bs f(\x'))$.
\begin{enumerate}
\item[{(C-1)}] Eigenfunctions $\{\phi_k(\bs y)\}_{k=1}^{\infty}$ of $g_*(\y,\y')$ defined in Theorem~\ref{theo:mercer} are continuously differentiable, i.e., $C^1$,
 and uniformly bounded in the sense of $\displaystyle \sup_{k \in \mathbb{N},\bs y \in \mathcal{Y}}|\phi_k(\bs y)| < \infty$
 and  $\displaystyle \sup_{k \in \mathbb{N},\bs y \in \mathcal{Y}}   \lambda_k  \| \partial \phi_k(\bs y)/\partial \bs y \|_2^2 < \infty$.
\item[{(C-2)}] $g_*(\bs y,\bs y')$ is $C^1$.
\item[{(C-3)}]  $\bs f_*$ is $C^1$.
\end{enumerate}

\textbf{NN architecture:} 
As we considered in Theorems~\ref{theo:universal_approximate} and \ref{theo:universal_cpd}, 
we employ a set of $K$-dimensional vector-valued NNs for $\mathcal{X}=[-M,M]^p$.
The activation function is confined to ReLU  $\sigma(z):=\max\{0,z\}$.
Let $L\in\mathbb{N}$ be the number of hidden layers, i.e., depth, of the NN, and
let $W\in\mathbb{N}$ be the total number of weights in the NN.
For example, $L=1$ and $W$ is the number of elements in $\bs A, \bs B, \bs c$ in Theorems~\ref{theo:universal_approximate}.
Instead of the fixed network architecture, here we consider a class of architectures specified by $W$ with a specific growing rate of the depth $L$.
For  $0\le \alpha \le 1$, define a set of all possible NNs with the constraint as
\begin{align}
	&\mathfrak{S}_\alpha(W,K)
	:=
	\{
	\bs f_{\text{NN}}:\mathcal{X} \to \mathbb{R}^K \mid 
    \bs f_{\text{NN}} \text{ has}
	\nonumber \\
	& \hspace{1em}
	\text{$W$ weights with depth $L=O((W/K)^\alpha)$}
	\},
	\label{eq:def_mlp}
\end{align}
where $W/K\to\infty$.
This is a simple extension of the case $K=1$ considered in \citet{yarotsky2018optimal},
where $\alpha=0$ and $\alpha=1$ correspond to constant-depth shallow NNs and constant-width deep NNs, respectively.

\begin{theo}[Approx. error rate for IPS]
\label{theo:universal_approximate_rate}
\normalfont 
Symbols and assumptions are the same as those of Theorem~\ref{theo:universal_approximate}
except for the additional conditions (C-1) and (C-2)
for $g_*^{\text{(PD)}}$ and (C-3) for $\bs f_*$.
Instead of the 1-hidden layer NN,
we consider the set of NNs $\bs f_{\text{NN}} \in \mathfrak{S}_\alpha(W_f,K)$
for $W_f \in \mathbb{N}$.
Then the approximation error rate of IPS is given by
\begin{align}
&
\scalebox{0.9}{$\displaystyle 
\inf_{\bs f_{\text{NN}} \in \mathfrak{S}_\alpha(W_f,K)}
\sup_{\bs x,\bs x' \in \mathcal{X}} $} \nonumber \\
& \hspace{3em}
\scalebox{0.9}{$\displaystyle 
    \bigg|
	g^{(\text{PD})}_*(\bs f_*(\bs x),\bs f_*(\bs x')) - \langle \bs f_{\text{NN}}(\bs x),\bs f_{\text{NN}}(\bs x') \rangle
\bigg| $}  \nonumber \\
&\hspace{1em}
\scalebox{0.9}{$\displaystyle 
=
O\Bigl(
	K^{-\tfrac{1}{K^*}}
	+
	K^{\tfrac{1}{2} + \tfrac{1+\alpha}{p}}
	W_f^{-\tfrac{1+\alpha}{p}}
\Bigr).
$} 
\label{eq:pd_error_bound}  
\end{align}
\end{theo}

Proof is in Supplement~\ref{sec:proof_universal_pd_rate}.
In the above result,
$O(K^{-1/K^*})$ is attributed to truncating (\ref{eq:mercer}) at $K$ terms
in Mercer's theorem
and $O(K^{\tfrac{1}{2} + \tfrac{1+\alpha}{p}} W_f^{-\tfrac{1+\alpha}{p}})$
is attributed to the approximation error of $\bs f_{\text{NN}}$.
The error rate for SIPS is similarly evaluated, but it includes the error rate for newly incorporated NN $u_{\text{NN}}$. 

\begin{theo}[Approx. error rate for SIPS]
\label{theo:universal_cpd_rate}
\normalfont
Symbols and assumptions are the same as those of Theorem~\ref{theo:universal_cpd}
except for the additional conditions
(C-1) for $g_0$ of (\ref{eq:g0cpd}),
(C-2) for $g^{(\text{CPD})}_*$, and
(C-3) for $\bs f_*$.
Instead of the 1-hidden layer NN,
we consider the set of NNs $\bs f_{\text{NN}} \in \mathfrak{S}_\alpha(W_f,K)$ for $W_f \in \mathbb{N}$ and
$u_{\text{NN}} \in \mathfrak{S}_\alpha(W_u,1)$ for $W_u \in \mathbb{N}$.
Then the approximation error rate of SIPS is given by
\begin{align}
&
\scalebox{0.9}{$\displaystyle 
\inf_{\substack{\bs f_{\text{NN}} \in \mathfrak{S}_\alpha(W_f,K) \\ u_{\text{NN}} \in \mathfrak{S}_\alpha(W_u,1)}}
\sup_{\bs x,\bs x' \in \mathcal{X}}
 \bigg|
	g^{(\text{CPD})}_*(\bs f_*(\bs x),\bs f_*(\bs x'))
 $} \nonumber \\
& \hspace{3em}
\scalebox{0.9}{$\displaystyle 
   - (\langle \bs f_{\text{NN}}(\bs x),\bs f_{\text{NN}}(\bs x') \rangle + u_{\text{NN}}(\bs x)+u_{\text{NN}}(\bs x'))
\bigg| $}  \nonumber \\
&\hspace{1em}
\scalebox{0.9}{$\displaystyle 
=
O\Bigl(
	K^{-\tfrac{1}{K^*}}
	+
	K^{\tfrac{1}{2} + \tfrac{1+\alpha}{p}}
	W_f^{-\tfrac{1+\alpha}{p}}
	+
	W_u^{-\tfrac{1+\alpha}{p}}
\Bigr).
$} 
\label{eq:cpd_error_bound}  
\end{align}
\end{theo}

Proof is in Supplement~\ref{sec:proof_universal_cpd_rate}.

In Theorems~\ref{theo:universal_approximate_rate} and \ref{theo:universal_cpd_rate},
the commonly appearing term $O(K^{-1/K^*})$ may be a bottleneck when $K^*$ is very large.
We may specify
$W_f = O(K^{1 + \tfrac{p}{1+\alpha}(\tfrac{1}{K^*} + \tfrac{1}{2})}) \approx O(K^{1 + \tfrac{p}{2(1+\alpha)}})$
and $W_u = O(K^{\tfrac{p}{(1+\alpha)K^*}})$ so that the overall approximation error rate is $O(K^{-1/K^*})$.

\begin{table*}[t]
\centering
\caption{Experiments on Co-authorship network and WordNet
 evaluated by ROC-AUC score (higher is better).
Sample average and the standard deviation of 5 runs are shown.}
% NOTE : 最低3回，目標は5回以上..とりあえず5回にしておく
%\vspace{0.5ex} % FIXME : これ使ってもいいのか？
\begin{adjustbox}{width=0.99\textwidth}
\begin{tabular}{lcccccccc}
%Similarity 
\toprule[0.2ex]
& \multicolumn{4}{c}{Co-authorship network} & \multicolumn{4}{c}{WordNet} \\\cmidrule(lr){2-5}\cmidrule(lr){6-9}
& $K=2$ & $K=5$ & $K=10$ & $K=20$ & $K=2$ & $K=5$ & $K=10$ & $K=20$ \\
\cmidrule(lr){2-5}\cmidrule(lr){6-9}

NSD & $\textbf{0.8220}\pm0.010$ & $0.8655\pm0.014$ & $0.8771\pm0.012$ & $0.8651\pm0.033$ & 
%$4168\pm3.835$ & $1924\pm3.104$ & $376.8\pm0.573$ & $13.64\pm0.163$ \\
$0.7924\pm0.0072$ & $0.8997\pm0.0009$ & $0.9569\pm0.0005$ & $0.9836\pm0.0001$ \\

Poincar\'e & $0.7071\pm0.021$ & $0.8738\pm0.001$ & $0.8822\pm0.001$ & $0.8835\pm0.001$ & 
%$\textbf{4.9}^{\dagger}$ & $4.02^{\dagger}$ & $3.84^{\dagger}$ & $3.98^{\dagger}$ \\
$0.8401\pm0.0073$ & $\textbf{0.9792}\pm0.0006$ & $\textbf{0.9866}\pm0.0003$ & $0.9851\pm0.0002$ \\

IPS & $0.7802\pm0.005$ & $0.8830\pm0.001$ & $0.8955\pm0.001$ & $0.8956\pm0.001$ & 
%$32.73\pm1.168$ & $4.037\pm0.383$ & $1.457\pm0.012$ & $1.271\pm0.070$ \\
$0.7245\pm0.0056$ & $0.7604\pm0.0055$ & $0.7688\pm0.0023$ & $0.7918\pm0.0018$ \\

\textbf{SIPS} & $0.7811\pm0.001$ & $\textbf{0.8853}\pm0.001$ & $\textbf{0.8964}\pm0.002$ & $\textbf{0.8974}\pm0.001$ & 
%$22.07\pm0.925$ & $\textbf{3.367}\pm0.111$ & $\textbf{1.439}\pm0.025$ & $\textbf{1.174}\pm0.006$ \\
$\textbf{0.9632}\pm0.0008$ & $0.9766\pm0.0006$  & $0.9825\pm0.0005$ & $\textbf{0.9865}\pm0.0004$ \\

\bottomrule[0.2ex]
\end{tabular}
\label{table:exp_results}
\end{adjustbox}
\end{table*}

\section{EXPERIMENTS}
In this section, we evaluate similarity models (NSD, Poincar\'e, IPS, SIPS)
on two real-world datasets:
Co-authorship network dataset~\citep{prado2013mining} in Section~\ref{sec:exp_coauthor}
and
WordNet dataset~\citep{miller1995wordnet} in Section~\ref{sec:exp_wordnet}. 
Details of experiments are shown in Supplement~\ref{sec:full_results_experiment}.

\subsection{Experiment on Co-authorship Network}
\label{sec:exp_coauthor}

Co-authorship network dataset \citep{prado2013mining} consists of $n=42,252$ nodes and $210,320$ undirected edges.
Each node $v_i$ represents an author, and data vector $\bs x_i \in \mathbb{R}^{33}$ ($p=33$) represents
the numbers of publications in $29$ conferences/journals and 
4 microscopic topological properties describing the direct neighborhood of the node.
Adjacency matrix $\bs W=(w_{ij}) \in \{0,1\}^{n \times n}$ represents the co-authorship relations: $w_{ij}=w_{ji}=1$ if $v_i$ and $v_j$ have any co-authorship relation, and $w_{ij}=w_{ji}=0$ otherwise.
% FIXME : preliminary-obtained data vectors の説明が詳しすぎるかも？　(いいけど場所が足りない)

\textbf{Preprocessing:} 
We split authors into training set (90\%) and test set (10\%).
Co-authorship relations for the test set are treated as unseen.
We use 10\% of the training set as validation set.

\textbf{Author feature vectors:} 
Using the data vectors for authors $\{\bs x_i\}_{i=1}^{n} \subset \mathbb{R}^p$, 
feature vectors $\{\bs y_i\}_{i=1}^{n} \subset \mathbb{R}^K$ are computed via a neural network $\bs y_i = \fnn(\bs x_i)$.
We employ $1$-hidden layer perceptron with $10,000$ hidden units and ReLU activation function.
For implementing SIPS, one of the $K$ output units of $\fnn(\bs x_i)$ is used for the bias term $u_i = u_\text{NN}(\bs x_i)$, so actually the feature vector is computed as $(\bs y_i, u_i) =  \fnn(\bs x_i) \in \mathbb{R}^K$ with $\bs y_i \in \mathbb{R}^{K-1}$.
%per each layer 
%The number of parameters is about $...$.
Model parameters are trained by maximizing 
the objective
\begin{align}
\sum_{1 \leq i \neq j \leq n}
w_{ij}
\log \frac{
\exp(h(\bs x_i,\bs x_j))
}{\sum_{k \in \mathcal{S}_r(\mathcal{N}_{ij})} \exp(h(\bs x_i,\bs x_k))}, 
\label{eq:objective}
\end{align}
where 
$h:\mathcal{X}^2 \to \mathbb{R}$ is a similarity function and
$\mathcal{S}_r(\mathcal{N}_{ij})$ is a subset that consists of $r=10$ entries randomly sampled from
$\mathcal{N}_{ij}:=\{k | 1 \leq k \leq n, w_{ik}=0\} \cup \{j\}$.
% NOTE (KIM) : \mid w_{ij}=0 はしていないのでコメントアウト

\textbf{Similarity models:}
(i)~NSD uses $h(\bs x_i,\bs x_j)=-\| \y_i - \y_j\|_2^2$.
(ii)~Poincar\'e embedding~\citep{nickel2017poincare} uses $h(\bs x_i,\bs x_j)=-d_{\text{Poincare}}(\y_i, \y_j)$ defined in (\ref{eq:dist_poincare}).
(iii)~IPS uses $h(\bs x_i,\bs x_j) =\langle \y_i, \y_j \rangle$.
(iv)~SIPS uses $h(\bs x_i,\bs x_j) =\langle \y_i, \y_j \rangle + u_i + u_j$.

\textbf{Results:} 
Models are evaluated by ROC-AUC~\citep{Bradley:1997:UAU:1746432.1746434}
on the task of predicting unseen co-authorship relations.
ROC-AUC scores are shown on the left-hand side of Table~\ref{table:exp_results}.
Although NSD demonstrates a good performance for $K=2$, SIPS outperforms the other methods for $K=5,10,20$.

\subsection{Experiment on WordNet}
\label{sec:exp_wordnet}

WordNet dataset \citep{miller1995wordnet} is a lexical resource that contains a variety of nouns and their relations. 
For instance, a noun ``mammal" represents a superordinate concept of a noun ``dog", thus these two words have hypernymy relation.
We preprocess WordNet dataset in the same way as \citet{nickel2017poincare}.
% so that $n=82,115$ nouns and $743,241$ hierarchical relations are extracted.
We used a subset of the graph with $n=4027$ nouns and $53,905$ hierarchical relations
by extracting all the nouns subordinate to ``animal''.
Each noun is represented by $v_i$, and relations are represented by adjacency matrix $\bs W=(w_{ij}) \in \{0,1\}^{n \times n}$, where  $w_{ij}=w_{ji}$ represents any hypernymy relation, including transitive closure,
between $v_i$ and $v_j$.

\textbf{Word feature vectors:}
Since nodes have no attributes, data vectors are formally treated as 1-hot vectors in $\mathbb{R}^n$.
Instead of learning neural networks, 
the distributed representations $\{\bs y_i\}_{i=1}^{n} \subset \mathbb{R}^K$ of words are learned by maximizing the objective (\ref{eq:objective}) with $r=20$ for NSD, Poincar\'e and IPS,
and $\{ (\y_i, u_i) \}_{i=1}^n \subset \mathbb{R}^K$ are learned for SIPS.
 Similarity models are the same as those of Section~\ref{sec:exp_coauthor}. 

\textbf{Results:} 
Models are evaluated by ROC-AUC of reconstruction error
on the task of reconstructing hierarchical relations in the same way as
\citet{nickel2017poincare}.
ROC-AUC score is listed on the right-hand side of Table~\ref{table:exp_results}. 
SIPS outperforms the other methods for $K=2, 20$,
and it is competitive to Poincar\'e embedding for $K=5, 10$.

\section{CONCLUSION}
We proposed a novel shifted inner-product similarity~(SIPS) for graph embedding~(GE), that is theoretically proved to approximate arbitrary conditionally positive-definite~(CPD) similarities including negative Poincar\'e distance.
Since SIPS automatically approximates a wide variety of similarities, SIPS alleviates the need for configuring the similarity function of GE.

\section*{Acknowledgement}
This work was partially supported by JSPS KAKENHI grant 16H02789 to HS and 17J03623 to AO.

%%% Referencesは無限ページ可能らしい(AISTATS2019)
%\bibliographystyle{apalike}
%\bibliography{representation}

\clearpage
\onecolumn

\begin{flushleft}
\textbf{\Large Supplementary Material:} \par
{\Large Graph Embedding with Shifted Inner Product Similarity and Its Improved Approximation Capability}
\end{flushleft}
\hrulefill

\appendix

\section{Experimental details}
\label{sec:full_results_experiment}

\textbf{Visualization of Fig.~\ref{fig:wordnet}:}
In Section~\ref{sec:exp_wordnet}, word feature vectors are computed from WordNet dataset.
We used feature vectors computed by SIPS with $K=5$.
Since $(\y_i, u_i)\in\mathbb{R}^5$ for SIPS,
we actually used $\y_i\in\mathbb{R}^4$ for the visualization.
We extracted 97 words from the $n=4027$ nouns, and applied t-SNE to $\{\y_i\}$ for the extracted words.
Words with any hypernymy relations are connected by segments.
In other words, $v_i$ and $v_j$ are connected when $w_{ij}=1$.
For extracting the 97 words, we chose the word ``animal'' as the root.
Then chose four subordinate words (``mammal", ``fish", ``reptile", ``invertebrate")  connected to the root,
and sampled more subordinate words from these four words, so that the total number of words becomes 97.
Words are grouped by the four subordinate words of the root, which are indicated by the colors.

\textbf{Optimization:}
In Section~\ref{sec:exp_coauthor}, all parameters are initialized as \citet{journals/corr/HeZR015} and trained by Adam~\citep{kingma2014adam} with initial learning rate 0.01 and batch size 64.
The number of iterations is 300,000.
To ensure robust comparison, we save model parameters at every 5,000 iterations, and select the best performance parameters tested on the validation set.
% Especially for Poincare embedding, we employ Riemannian SGD used in \citet{nickel2017poincare}
In Section~\ref{sec:exp_wordnet}, the most settings are the same as Section~\ref{sec:exp_coauthor}.
All parameters are initialized as \citet{journals/corr/HeZR015} and trained by Adam with initial learning rate 0.001 and batch size 128. The number of iterations is 150,000.

\section{Relationship between the Poisson model and the Bernoulli model}
\label{appendix:equivalence_poisson_logistic}

For a pair $(i,j) \in \mathcal{I}_n$, we consider the Poisson model
$ w_{ij} \sim \text{Po}(\lambda_{ij})$ with
$\lambda_{ij}=\exp(h(\bs x_i,\bs x_j))$.
In the below, $w_{ij}$ and $\lambda_{ij}$ are denoted as $w$ and $\lambda$ for simplifying the notation.
Noting $P(w=k) = \exp(-\lambda) \lambda^k/k!$ for $k \in \{0,1,\ldots,\}$, by Taylor expansion around $\lambda=0$, we have
$P(w=0) =  e^{-\lambda} = 1 - \lambda + \lambda^2/2 + O(\lambda^3)$
and
$P(w=1) = e^{-\lambda} \lambda = (1 - \lambda + O(\lambda^2))\lambda = \lambda - \lambda^2 + O(\lambda^3)$, and thus $P(w\ge2)=1-P(w=0)-P(w=1)=\lambda^2/2=O(\lambda^2)$.
On the other hand, $\sigma(h(\bs x_i,\bs x_j)) = (1+\lambda^{-1})^{-1} = \lambda - \lambda^2 + O(\lambda^3)$.
Therefore, $P(w=1) = \sigma(h(\bs x_i,\bs x_j)) + O(\lambda^3)$,
proving (\ref{eq:poisson-sigmoid}).

When link weights are very sparse as is often seen in applications, most of $\lambda_{ij}$'s will be very small. Then the above results imply that
$P(w_{ij}\ge2)\approx 0$ can be ignored and $P(w_{ij}=1) \approx \sigma(h(\bs x_i,\bs x_j))$ is interpreted as the Bernoulli model.

Let us consider a transformation from $w_{ij}$ to $\tilde{w}_{ij} \in \{0,1\}$ as
$\tilde{w}_{ij}:=\bs 1(w_{ij}>0)$.
By noting $P(\tilde w_{ij}=1) = P(w_{ij}>0) = 1 - P(w_{ij}=0) = \lambda_{ij} - \lambda_{ij}/2 + O(\lambda_{ij}^3)$, we have
\[
	P(\tilde{w}_{ij}  = 1 \mid \bs x_i,\bs x_j) = \sigma(h(\bs x_i,\bs x_j)) + O(\lambda_{ij}^2).
\]
Thus the Poisson model for $w_{ij}$ is also interpreted as
the Bernoulli model for the truncated variable $\tilde w_{ij}$.

\section{Proofs}
%%% このセクションはほぼワークショップ論文からのコピペ

\subsection{Proof of Proposition~\ref{prop:dist_approximation_lower_bound}}

\label{subsec:proof_lower_bound}

With $v=(2M)^{2p}$ and $\int=\int_{[-M,M]^p}$, a lower-bound of 
$\frac{1}{v}
\iint
|
	-\|\bs x-\bs x'\|_2^2
	-
	\langle
		\bs f(\bs x),\bs f(\bs x')
	\rangle
|
\diff \bs x
\diff \bs x' $ is derived as
\begin{align}
\frac{1}{v}
\iint
\bigg|
	-\|\bs x-\bs x'\|_2^2
	-
	\langle
		\bs f(\bs x),\bs f(\bs x')
	\rangle
\bigg|
\diff \bs x
\diff \bs x' 
&\geq
\bigg|
\frac{1}{v}
\iint
\left(
	-\|\bs x-\bs x'\|_2^2
	-
	\langle
		\bs f(\bs x),\bs f(\bs x')
	\rangle
\right)
\diff \bs x
\diff \bs x' \bigg| \nonumber \\
&=
\bigg|
\frac{1}{v}
\iint
\left(
	2\langle \bs x,\bs x' \rangle
	-
	\|\bs x\|_2^2
	-
	\|\bs x'\|_2^2
	-
	\langle
		\bs f(\bs x),\bs f(\bs x')
	\rangle
\right)
\diff \bs x
\diff \bs x' \bigg| \nonumber \\
&=
\bigg|
\frac{1}{v}
\left(
	2\bigg\| \int \bs x \diff \bs x\bigg\|_2^2
	-
	2 \int \diff \bs x \int \|\bs x\|_2^2 \diff \bs x
	-
	\bigg\|\int \bs f(\bs x) \diff \bs x \bigg\|_2^2
\right)
\bigg| \nonumber.
\end{align}
The terms in the last formula are computed as
$\int \bs x \diff \bs x=\bs 0,\int \diff \bs x
=
(2M)^p$,  %ここに全角「，」が入ってた！
\begin{align*}
\int \|\bs x\|_2^2\diff \bs x
&=
\sum_{i=1}^{p} \int x_i^2 \diff \bs x
=
(2M)^{p-1} \sum_{i=1}^{p} \int_{-M}^M x_i^2 \diff x_i
=
(2M)^{p-1}\frac{2pM^3}{3}
=
(2M)^p \frac{pM^2}{3}.
\end{align*}
Considering $\|\int f(\bs x)\diff \bs x\|_2^2 \geq 0$, we have
\begin{align*}
\frac{1}{v}
\iint
\bigg|
	-\|\bs x-\bs x'\|_2^2
	-
	\langle
		\bs f(\bs x),\bs f(\bs x')
	\rangle
\bigg|
\diff \bs x
\diff \bs x' 
\geq
\frac{2}{v}
\int \diff \bs x \int \|\bs x\|_2^2 \diff \bs x
=
\frac{2pM^2}{3}.
\end{align*}
Taking $\inf_{\bs f \in \mathfrak{S}(K)}$ proves the assertion.

\qed

\subsection{Proof of Theorem~\ref{theo:universal_cpd}~(Approximation theorem for SIPS)}
\label{sec:proof_universal_cpd}
Since $g^{\text{(CPD)}}_*:\mathcal{Y}^2 \to \mathbb{R}$ is a conditionally positive definite kernel on a compact set,
Lemma 2.1 of \citet{berg1984harmonic} indicates that 
$$
	g_{0}(\bs y_*,\bs y_*')
	:=
	g^{\text{(CPD)}}_*(\bs y_*,\bs y_*')
	-
	g^{\text{(CPD)}}_*(\bs y_*,\bs y_0)
	-
	g^{\text{(CPD)}}_*(\bs y_0,\bs y_*')
	+
	g^{\text{(CPD)}}_*(\bs y_0,\bs y_0)
$$
is positive definite for arbitrary $\bs y_0 \in \mathcal{Y}$. 
We fix $\bs y_0$ in the argument below. 
According to \citet{okuno2018probabilistic} Theorem~5.1 (Theorem~\ref{theo:universal_approximate} in this paper), we can specify a neural network $\bs f_{\text{NN}}(\bs x)$ such that
\[
	\sup_{\bs x,\bs x' \in \mathcal{X}}
	\bigg|
		g_0\left(\bs  f_*(\bs x),\bs f_*(\bs x') \right)
		-
		\langle
			\bs f_{\text{NN}}(\bs x)
			,
			\bs f_{\text{NN}}(\bs x')
		\rangle
	\bigg|
	<
	\varepsilon_1
\]
for any $\varepsilon_1$.
Next, let us consider a continuous function $h_*(\bs x):=g_*(\bs f_*(\bs x),\bs y_0)-\frac{1}{2}g_*(\bs y_0,\bs y_0)$. 
It follows from the universal approximation theorem \citep{cybenko1989approximation,pmlr-v70-telgarsky17a} that
for any $\varepsilon_2>0$, there exists $m_u \in \mathbb{N}$ such that
\[
	\sup_{\bs x \in \mathcal{X}} | h_*(\bs x)  - u_{\text{NN}}(\bs x)  | < \varepsilon_2.
\]
Therefore, we have
\begin{align}
\sup_{\bs x,\bs x' \in \mathcal{X}}
&\biggl|
g^{\text{(CPD)}}_*\left(
	\bs f_*(\bs x)
	,
	\bs f_*(\bs x')
\right)
-
\left\{
\langle \bs f_{\text{NN}}(\bs x),\bs f_{\text{NN}}(\bs x') \rangle
+
u_{\text{NN}}(\bs x)
+
u_{\text{NN}}(\bs x')
\right\}
\biggr| \nonumber \\
%%%%%%%%%%%%%%%%%%%%%%%%%%%%%%%%%%%%%%%%%%%%%%%%%%%%%%%%%%%
&=
\sup_{\bs x,\bs x' \in \mathcal{X}}
\biggl|
\left(
g_0\left(
	\bs f_*(\bs x)
	,
	\bs f_*(\bs x')
\right)
-
\langle \bs f_{\text{NN}}(\bs x),\bs  f_{\text{NN}}(\bs x') \rangle
\right) \nonumber \\
&\hspace{11em}
+
\left(
h_*(\bs x)
-
u_{\text{NN}}(\bs x)
\right)
+
\left(
h_*(\bs x')
-
u_{\text{NN}}(\bs x')
\right)
\biggr| \nonumber \\
%%%%%%%%%%%%%%%%%%%%%%%%%%%%%%%%%%%%%%%%%%%%%%%%%
&\leq 
\sup_{\bs x,\bs x' \in \mathcal{X}}
\biggl|
\left(
g_0\left(
	\bs f_*(\bs x)
	,
	\bs f_*(\bs x')
\right)
-
\langle \bs f_{\text{NN}}(\bs x), \bs f_{\text{NN}}(\bs x') \rangle
\right) \biggr| \nonumber \\
&\hspace{3em}
+
\sup_{\bs x \in \mathcal{X}}
\biggl|
h_*(\bs x)
-
u_{\text{NN}}(\bs x)
\biggr|
+
\sup_{\bs x' \in \mathcal{X}}
\biggl|
h_*(\bs x')
-
u_{\text{NN}}(\bs x')
\biggr| \label{eq:sips_inequality} \\
%%%%%%%%%%%%%%%%%%%%%%%%%%%%%%%%%%%%%%%%%%%%%%%%%%%%%%%%%%%
& <
\varepsilon_1 + 2 \varepsilon_2. \nonumber 
\end{align}
By letting
$\varepsilon_1=\varepsilon/2,
\varepsilon_2=\varepsilon/4$,
the last formula becomes smaller than $\varepsilon$, thus proving
\[
\sup_{\bs x,\bs x' \in \mathcal{X}}
\biggl|
g^{\text{(CPD)}}_*\left(
	\bs f_*(\bs x)
	,
	\bs f_*(\bs x')
\right)
-
\left\{
\langle \bs f_{\text{NN}}(\bs x),\bs f_{\text{NN}}(\bs x') \rangle
+
u_{\text{NN}}(\bs x)
+
u_{\text{NN}}(\bs x')
\right\}
\biggr|
<
\varepsilon.
\]

\qed

\subsection{Proof of Theorem~\ref{theo:universal_cpd2}~(Approximation theorem for C-SIPS)}
\label{sec:proof_universal_cpd2}
With fixed $\bs y_0 \in \mathcal{Y}$, it follows from \citet{berg1984harmonic} Lemma~2.1 and CPD-ness of the kernel $g^{\text{(CPD)}}_*$ that 
%$g_*$は$\mathcal{Y}^2$上CPDとし，
%$\bs y_0 \in \mathcal{Y}$を任意にとって固定する．
%\citet{berg1984harmonic} Lemma~2.1より
$$
	g_0(\bs y,\bs y')
	:=
	g^{\text{(CPD)}}_*(\bs y,\bs y')
	-
	g^{\text{(CPD)}}_*(\bs y,\bs y_0)
	-
	g^{\text{(CPD)}}_*(\bs y_0,\bs y')
	+
	g^{\text{(CPD)}}_*(\bs y_0,\bs y_0)
$$
is PD.
Since $\mathcal{Y}$ is compact, we have $\sup_{\bs y \in \mathcal{Y}}|g^{\text{(CPD)}}_*(\bs y,\bs y_0)|=a^2$ is bounded.
Let us take a sufficiently large $r>a$ and define  $\tau(\bs y):=\sqrt{r^2 + g^{\text{(CPD)}}_*(\bs y,\bs y_0)}$.
We consider a new kernel
\begin{align*}
g_1(\bs y,\bs y')
:=
g_0(\bs y,\bs y')
+
2\tau(\bs y)\tau(\bs y').
\end{align*}
Since both $g_0(\bs y,\bs y')$ and $\tau(\bs y)\tau(\bs y')$ are PD, $g_1(\bs y,\bs y')$ is also PD. 
Applying Taylor's expansion $\sqrt{1+x}=1+x/2+O(x^2)$, we have
%ここで，
%$\sup_{\bs y,\bs y' \in \mathcal{Y}}|g_*(\bs y,\bs y_0)|<r^2$となるように$r>0$を定める．
%$\tau(\bs y):=\sqrt{r^2 + g(\bs y,\bs y_0)}$とすると，単純なテイラー展開
%$\sqrt{1+x}=1+x/2+O(x^2)$により
\begin{align*}
\tau(\bs y)
\tau(\bs y')
&=
\sqrt{r^2+g^{\text{(CPD)}}_*(\bs y,\bs y_0)}
\sqrt{r^2+g^{\text{(CPD)}}_*(\bs y',\bs y_0)} \\
&=
r^2
\sqrt{1+g^{\text{(CPD)}}_*(\bs y,\bs y_0)/r^2}
\sqrt{1+g^{\text{(CPD)}}_*(\bs y',\bs y_0)/r^2} \\
&=
r^2
(
	1
	+
	g^{\text{(CPD)}}_*(\bs y,\bs y_0)/2r^2
	+
	O(r^{-4})	
)
(
	1
	+
	g^{\text{(CPD)}}_*(\bs y',\bs y_0)/2r^2
	+
	O(r^{-4})	
) \\
&=
r^2 + \frac{1}{2} (g^{\text{(CPD)}}_*(\bs y,\bs y_0) + g^{\text{(CPD)}}_*(\bs y',\bs y_0))
+
O(r^{-2}),
\end{align*}
thus proving
\begin{align*}
g_1(\bs y,\bs y')
=
g_0(\bs y,\bs y')
+
2\tau(\bs y)\tau(\bs y')
&=
g^{\text{(CPD)}}_*(\bs y,\bs y')
+
g^{\text{(CPD)}}_*(\bs y_0,\bs y_0)
+
2r^2 + O(r^{-2}). 
\end{align*}
Let us define $\gamma:=g^{\text{(CPD)}}_*(\bs y_0,\bs y_0)+2r^2=O(r^2)$. 
Considering the PD-ness of $g_1(\bs y,\bs y') = g^{\text{(CPD)}}_*(\bs y,\bs y') + \gamma+ O(r^{-2})$, we have
\begin{align}
&\sup_{\bs x,\bs x' \in \mathcal{X}}\bigg|
	g^{\text{(CPD)}}_*(\bs f_*(\bs x),\bs f_*(\bs x'))
	-
	\left(
		\langle \bs f_{\text{NN}}(\bs x),\bs f_{\text{NN}}(\bs x') \rangle
		-
		\gamma
	\right)
\bigg| \nonumber \\
&=
\sup_{\bs x,\bs x' \in \mathcal{X}}
\bigg|
	g_1(\bs f_*(\bs x),\bs f_*(\bs x'))
	-
	\langle \bs f_{\text{NN}}(\bs x), \bs f_{\text{NN}}(\bs x') \rangle
\bigg|
+
O(r^{-2}) \label{eq:csips_equality} \\
&<
\varepsilon + O(r^{-2}).\nonumber
\end{align}

\qed

\section{Approximation Error Rate}
We first discuss the approximation error rate for truncating the series expansion of
Mercer's theorem in Section~\ref{subsec:kernel_error_rate} and the approximation error rate for NNs in Section~\ref{subsec:nn_error_rate}. 
Then, by considering these error rates, we prove Theorems~\ref{theo:universal_approximate_rate} and \ref{theo:universal_cpd_rate} 
for IPS and SIPS, respectively,
in Sections~\ref{sec:proof_universal_pd_rate} and \ref{sec:proof_universal_cpd_rate}.

\subsection{Error rate for Mercer's theorem}
\label{subsec:kernel_error_rate}

We evaluate the error rate for Mercer's theorem~(shown as Theorem~\ref{theo:mercer} in this paper) to approximate PD kernels $g_*$ satisfying conditions (C-1) and (C-2) of Section~\ref{sec:error_bound_evaluation}.

We define the error rate for Mercer's theorem as
\begin{align}
\varepsilon_1(K):=\sup_{\bs y,\bs y' \in \mathcal{Y}}
\bigg|g_*(\bs y,\bs y')-
\sum_{k=1}^{K}\lambda_k \phi_k(\bs y)\phi_k(\bs y')
\bigg|. 
\label{eq:error_1}
\end{align}
Then, the error rate is given in the lemma below.

\begin{lem}
\normalfont
\label{theo:kernel_error_rate}
For compact set $\mathcal{Y} \subset \mathbb{R}^{K^*}$, $K^* \in \mathbb{N}$,
we consider a PD kernel $g_*:\mathcal{Y}^2 \to \mathbb{R}$ which 
satisfies conditions (C-1) and (C-2). 
Then, $\varepsilon_1(K)=O(K^{-1/K^*})$. 
\end{lem}

For proving the lemma, we first show a result of the decay rate for eigenvalues.
The theorem below is a special case of Theorem~4 of \citet{cobos1990eigenvalues} by assuming $\mu$ as Lebesgue measure, and $\bs \Omega=\mathcal{Y}$. 

%A special case of といってもほとんどもとのままなので
%pはかぶるのでLに変更
%alphaの値域修正
\begin{theo}[\citet{cobos1990eigenvalues}]
\normalfont
\label{prop:decay_eigenvalues}
Let $\mathcal{Y} \subset \mathbb{R}^L$ be a non-empty compact set for $L\in\mathbb{N}$, and let 
$g:\mathcal{Y}^2 \to \mathbb{R}$ be a positive definite kernel satisfying 
$\int_{\mathcal{Y}} 
\|g(\bs t,\cdot)\|_{C^{\alpha}}
\diff \bs t < \infty$, where $0<\alpha\le 1$ and 
\[
\|g(\bs t,\cdot)\|_{C^{\alpha}}
:=
\max\left\{\sup_{\bs y \in \mathcal{Y}}|g(\bs t,\bs y)|,\sup_{\substack{\bs y,\bs y' \in \mathcal{Y} \\ \bs y \neq \bs y'}}\frac{|g(\bs t,\bs y)-g(\bs t,\bs y')|}{\|\bs y-\bs y'\|_2^{\alpha}}\right\}. 
\]
Then, the $k$-th largest eigenvalue of $g$ is 
\[
	\lambda_k=O(k^{-1-\alpha/L}).
\]
\end{theo}

We apply Theorem~\ref{prop:decay_eigenvalues} to $g_*$ by letting $L=K^*$ and $\alpha=1$.
Then the eigenvalues of $g_*$ satisfy
\begin{align}
		\lambda_k=O(k^{-1-1/K^*}),
		\label{eq:decay_rate}
\end{align}
where the condition of $g$ in Theorem~\ref{prop:decay_eigenvalues} will be verified later.
On the other hand, Mercer's theorem and the condition (C-1) leads to
\begin{align}
\varepsilon_1(K)
&=
\sup_{\bs y,\bs y' \in \mathcal{Y}} \bigg| \sum_{k=K+1}^{\infty} \lambda_k \phi_k(\bs y)\phi_k(\bs y')\bigg| 
\leq
\sum_{k=K+1}^{\infty}
\lambda_k
\sup_{\bs y \in \mathcal{Y},l \in \mathbb{N}}|\phi_l(\bs y)|
\sup_{\bs y' \in \mathcal{Y},l' \in \mathbb{N}}|\phi_{l'}(\bs y')| \nonumber \\
&=
\left(\sup_{\bs y \in \mathcal{Y},k \in \mathbb{N}} |\phi_k(\bs y)| \right)^2
\sum_{k=K+1}^{\infty} \lambda_k
=
O\left( \sum_{k=K+1}^{\infty} \lambda_k \right). 
\label{eq:error_evaluation_1}
\end{align}
Therefore, substituting (\ref{eq:decay_rate}) into (\ref{eq:error_evaluation_1}), we have
\begin{align*}
		\varepsilon_1(K)
		=
		O\left(
		\sum_{k=K+1}^{\infty} \lambda_k
		\right)
		=
		O\left(
			\int_{K}^{\infty} k^{-1-1/K^*} \mathrm{d}k
		\right)
		=
		O\left(
			\bigg[-K^* k^{-1/K^*}\bigg]_{K}^{\infty}
		\right)
		=
		O(
			K^{-1/K^*}
		).
\end{align*}	
This proves Lemma~\ref{theo:kernel_error_rate}.
Finally, we verify that $g_*$ satisfies the condition of $g$ in Theorem~\ref{prop:decay_eigenvalues}.
As $g_*$ is continuous on compact set, 
	\begin{align}
	\sup_{\bs t \in \mathcal{Y}} \sup_{\bs y \in \mathcal{Y}} |g_*(\bs t,\bs y)|<\infty \label{eq:n1_bound}
	\end{align}
	obviously holds, and the condition (C-2) implies $\alpha$-H\"{o}lder continuity, and so
	\begin{align}		
		\sup_{\bs t \in \mathcal{Y}} \sup_{\substack{\bs y,\bs y' \in \mathcal{Y} \\ \bs y \neq \bs y'}}\frac{|g_*(\bs t,\bs y)-g_*(\bs t,\bs y')|}{\|\bs y-\bs y'\|_2}
		<
		\infty.
		\label{eq:n2_bound}
	\end{align}
Inequalities (\ref{eq:n1_bound}) and (\ref{eq:n2_bound}) lead to 
\begin{align*}
	\sup_{\bs t \in \mathcal{Y}}
	\|g_*(\bs t,\cdot)\|_{C^{1}}
	\leq
	\max\bigg\{
		\sup_{\bs t \in \mathcal{Y}} \sup_{\bs y \in \mathcal{Y}} |g_*(\bs t,\bs y)|
		,
		\sup_{\bs t \in \mathcal{Y}} \sup_{\substack{\bs y,\bs y' \in \mathcal{Y} \\ \bs y \neq \bs y'}}\frac{|g_*(\bs t,\bs y)-g_*(\bs t,\bs y')|}{\|\bs y-\bs y'\|_2}
	\bigg\} 
	<
	\infty.
\end{align*}		
	Thus $g_*$ satisfies
	\begin{align*}
		\int_{\mathcal{Y}} \|g_*(\bs t,\cdot)\|_{C^{1}} \diff \bs t
		&\leq
		\sup_{\bs t \in \mathcal{Y}}
		\|g_*(\bs t,\cdot)\|_{C^{1}}
		\int_{\mathcal{Y}} \diff \bs t
		<
		\infty,
	\end{align*}
because compact set $\mathcal{Y} \subset \mathbb{R}^{K^*}$ is bounded and closed. 
\qed

\subsection{Error rate for NN approximations}
\label{subsec:nn_error_rate}

We refer to the result of \citet{yarotsky2018optimal}.
By combining Proposition~1 ($\alpha=0$, i.e., constant-depth shallow NNs) and Theorem~2 ($0<\alpha\le 1$, i.e., deep NNs with growing depth as $W$ increases) of \citet{yarotsky2018optimal}, we have the following theorem.

\begin{theo}[\citet{yarotsky2018optimal}]
\normalfont
\label{theo:yarotsky2018}
For $\mathcal{X}=[-M,M]^p$, $M>0$, $p \in \mathbb{N}$ and $0\le \alpha \le 1$,
we consider the set of real-valued NNs $v_{\text{NN}} \in \mathfrak{S}_\alpha(W,1)$ for $W\in\mathbb{N}$.
Let $\omega(v;r):=\max\{|v(\bs x)-v(\bs x')| : \bs x,\bs x' \in \mathcal{X}, \|\bs x-\bs x'\| \leq r \}$ be the modulus of continuity.
Then, there exist $a,c \in \mathbb{R}$ such that
\begin{align*}
\inf_{v_{\text{NN}} \in \mathfrak{S}_\alpha(W,1)}
\sup_{\bs x \in \mathcal{X}}
|v_*(\bs x)-v_{\text{NN}}(\bs x)|
\leq 
a \omega(v_*;c W^{-\tfrac{1+\alpha}{p}})
\end{align*}
holds for any real-valued continuous function $v_*:\mathcal{X} \to \mathbb{R}$.
\end{theo}

In later sections, we will use the following two lemmas, which are immediate consequences of Theorem~\ref{theo:yarotsky2018}.

\begin{lem}
\label{lem:nn_approximation_rate}
\normalfont 
Symbols are the same as those of Theorem~\ref{theo:yarotsky2018}.
Assume that $v_*$ is continuously differentiable over $\mathcal{X}$,
and fix such a $v^*$.
Then, as $W\to\infty$,  we have
\begin{align*}
\inf_{v_{\text{NN}} \in \mathfrak{S}_\alpha(W,1)}
\sup_{\bs x \in \mathcal{X}}
|v_*(\bs x)-v_{\text{NN}}(\bs x)|
=
O(W^{-\tfrac{1+\alpha}{p}}).
\end{align*}
\end{lem}

Proof is based on the intermediate value theorem. 
For $\bs x,\bs x' \in \mathcal{X}$ satisfying $\|\bs x-\bs x'\| \leq r$, there exists $\bs x_0 \in \mathcal{X}$ such that 
$v_*(\bs x)-v_*(\bs x')=\frac{\partial v_*(\bs x)}{\partial \bs x} |_{\bs x=\bs x_0}(\bs x-\bs x')$. 
Since $b:=\sup_{\bs x \in \mathcal{X}}\| \partial v_*(\bs x)/\partial \bs x\|$ is bounded because of the continuity of the first-order derivative $\partial v_*(\bs x)/\partial \bs x$, 
Cauchy-Schwarz inequality indicates 
\begin{align*}
|v_*(\bs x)-v_*(\bs x')| \leq \bigg\|\frac{\partial v_*(\bs x)}{\partial \bs x} \bigg|_{\bs x=\bs x_0}\bigg\|_2 \|\bs x-\bs x'\|_2
\leq 
br.
\end{align*} 
Thus we have $\omega(v_*;r) \leq br$, indicating 
\begin{align}
a\omega(v_*;cW^{-\tfrac{1+\alpha}{p}}) \leq abcW^{-\tfrac{1+\alpha}{p}}. \label{eq:omega_bound}
\end{align}
Substituting (\ref{eq:omega_bound}) into Theorem~\ref{theo:yarotsky2018} proves the lemma. \qed

\begin{lem}
\label{lem:nn_approximation_rate_vector}
\normalfont 
For $\mathcal{X}=[-M,M]^p$, $M>0$, $p \in \mathbb{N}$ and $0\le \alpha \le 1$,
we consider the set of NNs $\bs v_{\text{NN}} \in \mathfrak{S}_\alpha(W,K)$ for $W, K\in\mathbb{N}$.
Let $\bs v_*:\mathcal{X} \to \mathbb{R}^K$ be a vector-valued continuously differentiable function over $\mathcal{X}$
such that
$\sup_{k\in \{1,\ldots,K\}, \x \in \mathcal{X}}  \|\partial v_{*k}(\x) / \partial \x\|_2 \le b$ for some $b$ which does not depend on $K$.
Then, as $W/K\to\infty$, we have
\begin{align*}
\inf_{\bs v_{\text{NN}} \in \mathfrak{S}_\alpha(W,K)}
\sup_{\bs x \in \mathcal{X}}
\|\bs v_*(\bs x)-\bs v_{\text{NN}}(\bs x)\|_2
=
O(K^{\tfrac{1}{2} + \tfrac{1+\alpha}{p}} W^{-\tfrac{1+\alpha}{p}}).
\end{align*}
\end{lem}

Proof is based on applying Lemma~\ref{lem:nn_approximation_rate} to each of $K$ output units of $\bs v_*$.
\begin{figure}[htbp]
\centering
	\includegraphics[width=15cm]{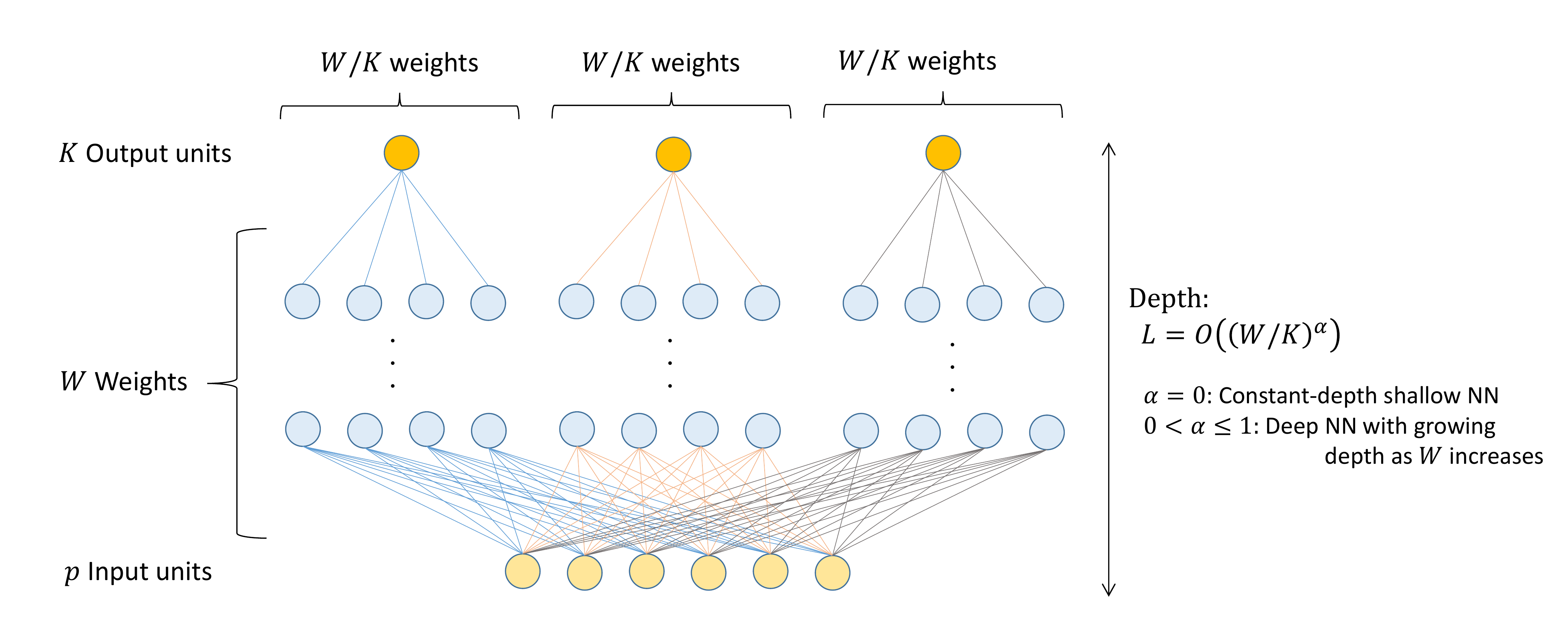}
	\caption{A structure of vector-valued neural network $\bs v_{\text{NN}}:\mathbb{R}^p \to \mathbb{R}^K$ having $W$ weights. We allocate $W/K$ weights to each output unit, so that weights are not shared by the $K$ output units.
	In practice, internal units are often shared by the output units, but we consider the above structure for showing 
	the upper bound of the approximation error.
	} 
	\label{fig:vvnn}
\end{figure}
We consider $K$ real-valued neural networks of depth $L=O((W/K)^\alpha)$ with $W/K$ weights as shown in Fig.~\ref{fig:vvnn}.
Since such NNs are included in $\mathfrak{S}_\alpha(W,K)$,
we have
\begin{align*}
\inf_{\bs v_{\text{NN}} \in \mathfrak{S}_\alpha(W,K)}
\sup_{\bs x \in \mathcal{X}}
\|\bs v_*(\bs x))-\bs v_{\text{NN}}(\bs x)\|_2
&\le
\Bigl(
\sum_{k=1}^{K} 
\inf_{v_k \in \mathfrak{S}_\alpha(W/K,1)}
\sup_{\bs x \in \mathcal{X}}
   |v_{*k}(\bs x)-v_k(\bs x) |^2
\Bigr)^{1/2},
\end{align*}
where $\bs v_*(\bs x)=(v_{*1}(\bs x),v_{*2}(\bs x),\ldots,v_{*K}(\bs x)),
\bs v_{\text{NN}}(\bs x)=(v_1(\bs x),v_2(\bs x),\ldots,v_K(\bs x))$. 
We apply Lemma~\ref{lem:nn_approximation_rate} with $W/K$ weights 
to each $v_{*k}$, where the same bound $b$ is used in (\ref{eq:omega_bound}).
Then the error is bounded by $\sqrt{K} \times abc (W/K)^{-\tfrac{1+\alpha}{p}} =
O(K^{\tfrac{1}{2} + \tfrac{1+\alpha}{p}} W^{-\tfrac{1+\alpha}{p}}) $.
\qed

\subsection{Proof of Theorem~\ref{theo:universal_approximate_rate}~(Approximation error rate for IPS)}
\label{sec:proof_universal_pd_rate}

Applying Theorem~\ref{theo:mercer} to a PD kernel $g^{\text{(PD)}}_*$,
there exist eigenvalues  $\{\lambda_k\}_{k=1}^{\infty}$, $\lambda_1\ge \lambda_2 \ge \cdots$ and
eigenfunctions $\{\phi_k(\bs y)\}_{k=1}^{\infty}$  such that
$\sum_{k=1}^{K}\lambda_k \phi_k(\bs y)\phi_k(\bs y')$ absolutely and uniformly converges to $g^{\text{(PD)}}_*(\bs y,\bs y')$ as $K \to \infty$. 
Here, we define two vector-valued functions 
\begin{align*}
	\bs \eta_K(\bs y)
	&:=
	(\lambda_1^{1/2}\phi_1(\bs y),
	\lambda_2^{1/2}\phi_2(\bs y),
\ldots,
	\lambda_K^{1/2}\phi_K(\bs y)), \\
	\tilde{\bs \phi}_K(\bs x)
	&:=
	\bs \eta_K(\bs f_*(\bs x)), 
\end{align*}
so that $\displaystyle \langle \bs \eta_K(\bs f_*(\bs x)) ,\bs \eta_K(\bs f_*(\bs x')) \rangle = \langle \tilde{\bs \phi}_K(\bs x),\tilde{\bs \phi}_K(\bs x') \rangle=\sum_{k=1}^{K}\lambda_k \phi_k(\bs f_*(\bs x))\phi_k(\bs f_*(\bs x'))$. 
Using these functions, for any $\bs f_{\text{NN}} \in \mathfrak{S}_\alpha(W_f,K)$, we have
\begin{align}
&
	\bigg|
		g^{\text{(PD)}}_*(\bs f_*(\bs x),\bs f_*(\bs x'))
		-
		\langle \bs f_{\text{NN}}(\bs x),\bs f_{\text{NN}}(\bs x') \rangle
	\bigg| \nonumber \\
&\hspace{3em}\leq
	\bigg|
		g_*(\bs f_*(\bs x),\bs f_*(\bs x'))
		-
		\langle \bs \eta_K(\bs f_*(\bs x)),\bs \eta_K(\bs f_*(\bs x')) \rangle
	\bigg| \label{eq:pf1_term1} \\
&\hspace{6em}+
	\bigg|
		\langle \tilde{\bs \phi}_K(\bs x),\tilde{\bs \phi}_K(\bs x') \rangle
		-
		\langle \tilde{\bs \phi}_K(\bs x),\bs f_{\text{NN}}(\bs x') \rangle
	\bigg| 
+
	\bigg|
		\langle \tilde{\bs \phi}_K(\bs x),\bs f_{\text{NN}}(\bs x') \rangle
		-
		\langle \bs f_{\text{NN}}(\bs x),\bs f_{\text{NN}}(\bs x') \rangle
	\bigg|. \label{eq:pf1_term2}	
\end{align}
These terms (\ref{eq:pf1_term1}) and (\ref{eq:pf1_term2}) can be evaluated in the following way. 
\begin{itemize}
\item Regarding the term (\ref{eq:pf1_term1}), 
\begin{align*}
&\sup_{\bs x,\bs x' \in \mathcal{X}}\bigg|
		g^{\text{(PD)}}_*(\bs f_*(\bs x),\bs f_*(\bs x'))
		-
		\langle \bs \eta_K(\bs f_*(\bs x)),\bs \eta_K(\bs f_*(\bs x')) \rangle
	\bigg| \\
	&\hspace{2em}\leq
	\sup_{\bs y,\bs y' \in \mathcal{Y}}\bigg|
		g^{\text{(PD)}}_*(\bs y,\bs y')
		-
		\langle \bs \eta_K(\bs y),\bs \eta_K(\bs y') \rangle
	\bigg| \\
	&\hspace{2em} =
	\sup_{\bs y,\bs y' \in \mathcal{Y}}\bigg|
		g^{\text{(PD)}}_*(\bs y,\bs y')
		-
		\sum_{k=1}^{K}\lambda_k \phi_k(\bs y)\phi_k(\bs y')
	\bigg| = O(K^{-1/K^*}),
\end{align*}
where the last formula follows by applying  Lemma~\ref{theo:kernel_error_rate} to $g^{\text{(PD)}}_*$.
Thus, we have 
\begin{align}
\inf_{\bs f_{\text{NN}} \in \mathfrak{S}_\alpha(W_f,K)}\sup_{\bs x,\bs x' \in \mathcal{X}}\bigg|
		g^{\text{(PD)}}_*(\bs f_*(\bs x),\bs f_*(\bs x'))
		-
		\langle \bs \eta_K(\bs f_*(\bs x)),\bs \eta_K(\bs f_*(\bs x')) \rangle \bigg|
		=
		O(K^{-1/K^*}).
		\label{eq:pf1_term1_rate}
		\end{align}

\item Regarding the term (\ref{eq:pf1_term2}), 
\begin{align}
	&\sup_{\bs x,\bs x' \in \mathcal{X}} \biggl\{ \bigg|
		\langle \tilde{\bs \phi}_K(\bs x),\tilde{\bs \phi}_K(\bs x') \rangle
		-
		\langle \tilde{\bs \phi}_K(\bs x),\bs f_{\text{NN}}(\bs x') \rangle
	\bigg| 
+
	\bigg|
		\langle \tilde{\bs \phi}_K(\bs x),\bs f_{\text{NN}}(\bs x') \rangle
		-
		\langle \bs f_{\text{NN}}(\bs x),\bs f_{\text{NN}}(\bs x') \rangle
	\bigg|  \biggr\}\nonumber \\
%%%======================================
	&\leq \sup_{\bs x,\bs x' \in \mathcal{X}} \biggl\{
	\| \tilde{\bs \phi}_K(\bs x) \|_2
	\| \tilde{\bs \phi}_K(\bs x')-\bs f_{\text{NN}}(\bs x')\|_2 
+
	\| \bs f_{\text{NN}}(\bs x') \|_2
	\| \tilde{\bs \phi}_K(\bs x)-\bs f_{\text{NN}}(\bs x)\|_2   \biggr\}\nonumber \\
%%%========================================
	&\leq \sup_{\bs x,\bs x' \in \mathcal{X}} \biggl\{
	\| \tilde{\bs \phi}_K(\bs x) \|_2
	\| \tilde{\bs \phi}_K(\bs x')-\bs f_{\text{NN}}(\bs x')\|_2 
+
	(\|\tilde{\bs \phi}_K(\bs x')\|_2
	+
	\| \tilde{\bs \phi}_K(\bs x')-\bs f_{\text{NN}}(\bs x') \|_2)
	\| \tilde{\bs \phi}_K(\bs x)-\bs f_{\text{NN}}(\bs x)\|_2   \biggr\}\nonumber \\
%%%========================================
	&=
2 \sup_{\bs x \in \mathcal{X}}  \|\tilde{\bs \phi}_K(\bs x)\|_2
  \sup_{\bs x' \in \mathcal{X}}   \| \tilde{\bs \phi}_K(\bs x')-\bs f_{\text{NN}}(\bs x')\|_2 
+ \sup_{\bs x \in \mathcal{X}} \| \tilde{\bs \phi}_K(\bs x)-\bs f_{\text{NN}}(\bs x) \|_2^2.
\nonumber
\end{align}
Here, $\|\tilde{\bs \phi}_K(\bs x)\|_2=\|\sum_{k=1}^{K} \lambda_k \phi_k(\bs f_*(\bs x))\phi_k(\bs f_*(\bs x))\|_2 \leq \|\sum_{k=1}^{\infty}\lambda_k \phi_k(\bs f_*(\bs x))\phi_k(\bs f_*(\bs x))\|_2=\|g^{\text{(PD)}}_*(\bs f_*(\bs x),\bs f_*(\bs x))\|_2$ is bounded, because $g^{\text{(PD)}}_*(\bs f_*(\bs x),\bs f_*(\bs x))$ is continuous over the compact set $\mathcal{X}^2$.
For applying Lemma~\ref{lem:nn_approximation_rate_vector} to $\tilde{\bs \phi}_K(\bs x)$, we need to show that the constant $b$ exists.
Noting
$\| \partial \tilde \phi_k/\partial \x\|_2^2 = \sum_{i=1}^p (\partial \tilde \phi_k/\partial x_i)^2
\le \sum_{i=1}^p \lambda_k \|\partial \phi_k/\partial \bs y\|_2^2 \,\|\partial \bs f_*/\partial x_i\|_2^2$,
we have
\begin{equation} \label{eq:difphibound}
	\sup_{k\in\mathbb{N}} \sup_{\x\in\mathcal{X}} \| \partial \tilde \phi_k/\partial \x\|_2^2 
	\le \sup_{k\in\mathbb{N}} \sup_{\bs y\in\mathcal{Y}} \lambda_k \|\partial \phi_k/\partial \bs y\|_2^2\,
	\sup_{\x\in\mathcal{X}} \sum_{i=1}^p \|\partial \bs f_*/\partial x_i\|_2^2 <\infty,
\end{equation}
where $\sup_{k\in\mathbb{N}} \sup_{\bs y\in\mathcal{Y}} \lambda_k \|\partial \phi_k/\partial \bs y\|_2^2<\infty$ follows from (C-1) and $\sup_{\x\in\mathcal{X}} \sum_{i=1}^p \|\partial \bs f_*/\partial x_i\|_2^2 <\infty$ follows from (C-3).
We can take $b^2$ as the upper bound of (\ref{eq:difphibound}), and then Lemma~\ref{lem:nn_approximation_rate_vector} implies
\begin{equation}
	\inf_{\bs f_{\text{NN}} \in \mathfrak{S}_\alpha(W_f,K)}\sup_{\bs x \in \mathcal{X}}
\| \tilde{\bs \phi}_K(\bs x)-\bs f_{\text{NN}}(\bs x)\|_2
=
O(K^{\tfrac{1}{2} + \tfrac{1+\alpha}{p}} W_f^{-\tfrac{1+\alpha}{p}})
\end{equation}
so that the evaluation of (\ref{eq:pf1_term2}) leads to
\begin{align}
&\scalebox{0.9}{$\displaystyle
\inf_{\bs f_{\text{NN}} \in \mathfrak{S}_\alpha(W_f,K)}
\sup_{\bs x,\bs x' \in \mathcal{X}}
\left\{
	\bigg|
		\langle \tilde{\bs \phi}_K(\bs x),\tilde{\bs \phi}_K(\bs x') \rangle
		-
		\langle \tilde{\bs \phi}_K(\bs x),\bs f_{\text{NN}}(\bs x') \rangle
	\bigg| 
+
	\bigg|
		\langle \tilde{\bs \phi}_K(\bs x),\bs f_{\text{NN}}(\bs x') \rangle
		-
		\langle \bs f_{\text{NN}}(\bs x),\bs f_{\text{NN}}(\bs x') \rangle
	\bigg|
\right\}
$} \nonumber \\
&\hspace{3em}=
O(K^{\tfrac{1}{2} + \tfrac{1+\alpha}{p}} W_f^{-\tfrac{1+\alpha}{p}}).
\label{eq:pf1_term2_rate}
\end{align}

\end{itemize}
Considering (\ref{eq:pf1_term1_rate}) and (\ref{eq:pf1_term2_rate}), we finally obtain
\begin{align*}
\inf_{\bs f_{\text{NN}} \in \mathfrak{S}_\sigma(W_f,K)}
\sup_{\bs x,\bs x' \in \mathcal{X}}
	\bigg|
		g^{\text{(PD)}}_*(\bs f_*(\bs x),\bs f_*(\bs x'))
		-
		\langle \bs f_{\text{NN}}(\bs x),\bs f_{\text{NN}}(\bs x') \rangle
	\bigg|
	=
	O\Bigl(
	K^{-\tfrac{1}{K^*}}+    K^{\tfrac{1}{2} + \tfrac{1+\alpha}{p}} W_f^{-\tfrac{1+\alpha}{p}}   
	\Bigr).
\end{align*}
\qed

\subsection{Proof of Theorem~\ref{theo:universal_cpd_rate}~(Approximation error rate for SIPS)}
\label{sec:proof_universal_cpd_rate}
%
%We first define a function
%\begin{align*}
%h_*(\bs x)
%:=
%g_*^{\text{(CPD)}}(\bs f_*(\bs x),\bs y_0)
%-
%\frac{1}{2}g_*^{\text{(CPD)}}(\bs y_0,\bs y_0),
%\end{align*}
%so that $g_*^{\text{(CPD)}}(\bs f_*(\bs x),\bs f_*(\bs x'))
%=
%g_0(\bs f_*(\bs x),\bs f_*(\bs x'))
%+
%h_*(\bs x)
%+
%h_*(\bs x')$, and $h_*(\bs x)$ is twice differentiable. 

Recall the inequality (\ref{eq:sips_inequality}) in Section~\ref{sec:proof_universal_cpd}.
\begin{align}
	&
	\sup_{\bs x,\bs x' \in \mathcal{X}}
	\bigg|
		g^{\text{(CPD)}}_*(\bs f_*(\bs x),\bs f_*(\bs x'))
		-
		(\langle \bs f_{\text{NN}}(\bs x),\bs f_{\text{NN}}(\bs x') \rangle+u_{\text{NN}}(\bs x)+u_{\text{NN}}(\bs x'))
	\bigg| \nonumber \\
%%=================================
	&\hspace{2em} \leq
	\sup_{\bs x,\bs x' \in \mathcal{X}}
	\bigg|
	g_0(\bs f_*(\bs x),\bs f_*(\bs x'))-\langle \bs f_{\text{NN}}(\bs x),\bs f_{\text{NN}}(\bs x') \rangle
	\bigg| 
	+
	2 \sup_{\bs x\in \mathcal{X}}
	\bigg|h_*(\bs x)-u_{\text{NN}}(\bs x)\bigg|
	\label{eq:pf2_bound}
\end{align}
We evaluate the two terms in (\ref{eq:pf2_bound}). 
Since we have assumed that $g^{\text{(CPD)}}_*$ is $C^1$ (the condition C-2), $g_0$ and $h_*$ are also $C^1$.
Then, by applying Theorem~\ref{theo:universal_approximate_rate} to the PD kernel $g_0$,
the first term in (\ref{eq:pf2_bound}) is evaluated as
\begin{align}
	\inf_{\bs f_{\text{NN}} \in \mathfrak{S}_\alpha(W_f,K)}\sup_{\bs x,\bs x' \in \mathcal{X}}
	\bigg|
	g_0(\bs f_*(\bs x),\bs f_*(\bs x'))-\langle \bs f_{\text{NN}}(\bs x),\bs f_{\text{NN}}(\bs x') \rangle
	\bigg| 
	=
	O\Bigl(K^{-\tfrac{1}{K^*}}
	+
	K^{\tfrac{1}{2} + \tfrac{1+\alpha}{p}}
	W_f^{-\tfrac{1+\alpha}{p}}\Bigr).
\label{eq:pf2_term1}
\end{align} 
By applying Lemma~\ref{lem:nn_approximation_rate} to $h_*$,
the second term in (\ref{eq:pf2_bound}) is evaluated as
\begin{align}
\inf_{u_{\text{NN}} \in \mathfrak{S}_\alpha(W_u,1)}\sup_{\bs x \in \mathcal{X}}
	\bigg|h_*(\bs x)-u_{\text{NN}}(\bs x)\bigg|
=
O\Bigl(W_u^{-\tfrac{1+\alpha}{p}}\Bigr). \label{eq:pf2_term2}
\end{align}
Considering (\ref{eq:pf2_bound}), (\ref{eq:pf2_term1}) and (\ref{eq:pf2_term2}), we obtain 
\begin{align*}
&\inf_{\substack{\bs f_{\text{NN}} \in \mathfrak{S}_\alpha(W_f,K) \\ u_{\text{NN}} \in \mathfrak{S}_\alpha(W_u,1)}}\sup_{\bs x,\bs x' \in \mathcal{X}}
\bigg|
		g^{\text{(CPD)}}_*(\bs f_*(\bs x),\bs f_*(\bs x'))
		-
		(\langle \bs f_{\text{NN}}(\bs x),\bs f_{\text{NN}}(\bs x') \rangle+u_{\text{NN}}(\bs x)+u_{\text{NN}}(\bs x'))
	\bigg| \\
	&\hspace{3em}
=
O\Bigl(
	K^{-\tfrac{1}{K^*}}
	+
	K^{\tfrac{1}{2} + \tfrac{1+\alpha}{p}}
	W_f^{-\tfrac{1+\alpha}{p}}
	+
	W_u^{-\tfrac{1+\alpha}{p}}
\Bigr).
\end{align*}
\qed

%%% ワークショップ論文からのコピペ
\section{Non-CPD Similarities}
\label{appendix:non_cpd}
CPD includes a broad range of kernels, but there exists a variety of non-CPD kernels. 
One example is Epanechnikov kernel $g(\bs y,\bs y'):=(1-\|\bs y-\bs y'\|_2^2)\bs 1(\|\bs y-\bs y'\|_2 \leq 1)$. 
To approximate similarities based on such non-CPD kernels, we propose a novel model, yet based on inner product, with high approximation capability beyond SIPS.
Although parameter optimization of this model is not always easy due to the excessive degrees of freedom,
the model is, in theory, shown to be capable of approximating more general kernels that are considered in \citet{ong2004learning}.

\subsection{Proposed model}

Let us consider a similarity $h(\bs x,\bs x')=g_*(f_*(\bs x),f_*(\bs x'))$ with any kernel $g_*:\mathbb{R}^{2K^*} \to \mathbb{R}$ and a continuous map $f_*:\mathbb{R}^p \to \mathbb{R}^{K^*}$.
To approximate it, we consider a similarity model  % さすがにnewとは言いにくい
\begin{align}
h(\bs x_i, \bs x_j) = 
\langle \bs f_{\text{NN}}(\bs x_i),\bs f_{\text{NN}}(\bs x_j) \rangle - \langle \bs r_{\text{NN}}(\bs x_i),\bs r_{\text{NN}}(\bs x_j) \rangle,
\label{eq:general_ips}
\end{align}
where $\bs f_{\text{NN}}:\mathbb{R}^{p} \to \mathbb{R}^{K_+}$ and $\bs r_{\text{NN}}:\mathbb{R}^p \to \mathbb{R}^{K_-}$ are neural networks. 
Since the kernel $g(\bs y, \bs y') = \langle \bs y_+,\bs y'_+ \rangle - \langle \bs y_-,\bs y'_- \rangle$ with respect to $\bs y = (\bs y_+,\bs y_-) \in \mathbb{R}^{K_+ + K_-}$ represents the difference of two IPSs, we call (\ref{eq:general_ips}) as inner product difference similarity~(IPDS) model.

By replacing $\bs f_{\text{NN}}(\bs x)$ and $\bs r_{\text{NN}}(\bs x)$ with 
$(\bs f_{\text{NN}}(\bs x)^{\top},u_{\text{NN}}(\bs x),1)^{\top}$ and $u_{\text{NN}}(\bs x)-1 \in \mathbb{R}$, respectively, IPDS reduces to SIPS defined in eq.~(\ref{eq:sips}), meaning that IPDS includes SIPS as a special case. 
Therefore, IPDS approximates any CPD similarities arbitrary well. 
Further, we prove that IPDS approximates more general similarities arbitrary well.

\subsection{Approximation theorem}
\begin{theo}[Approximation theorem for IPDS]
\normalfont
\label{theo:universal_general}
Symbols and assumptions are the same as those of Theorem~\ref{theo:universal_cpd} but $g_*$ is a general kernel, which is only required to be dominated by some PD kernels $g$, i.e., $g-g_*$ is PD.
%記号はTheorem~\ref{theo:universal_cpd}と同様とする．
For arbitrary $\varepsilon>0$, 
by specifying sufficiently large $K_+, K_- \in \mathbb{N},m_+=m_+(K_+),m_-=m_-(K_-) \in \mathbb{N}$, 
there exist $\bs A \in \mathbb{R}^{K_+ \times m_+},\bs B \in \mathbb{R}^{m_+ \times p},\bs c \in \mathbb{R}^{m_+},
\bs E \in \mathbb{R}^{K_- \times m_-}, \bs F \in \mathbb{R}^{m_- \times p}, \bs o \in \mathbb{R}^{m_-}$ such that
\begin{align*}
&\scalebox{0.95}{$
\bigg|
g_*\left(f_*(\x),f_*(\x')\right)
-
\left(\big\langle \bs f_{\text{NN}}(\x), \bs f_{\text{NN}}(\x') \big\rangle
-\big \langle \bs r_{\text{NN}}(\x),\bs r_{\text{NN}}(\x') \big \rangle
\right)
\bigg|
<\varepsilon$}
\end{align*}
for all $(\bs x,\bs x') \in [-M,M]^{2p}$, where $\bs f_{\text{NN}}(\x)
=
\bs A \bs \sigma(\bs B\x + \bs c) \in \mathbb{R}^{K_+}$ and
$\bs r_{\text{NN}}(x)
=
\bs E \bs \sigma(\bs F\x + \bs o) \in \mathbb{R}^{K_-}$ are 1-hidden layer neural networks with $m_+$ and $m_-$ hidden units, respectively.
\end{theo}

In theorem~\ref{theo:universal_general}, the kernel $g_*$ is only required to be dominated by some PD kernels, thus $g_*$ is not limited to CPD. 
We call such a kernel $g_*$ satisfying the condition in Theorem~\ref{theo:universal_general}, i.e., there exists a PD kernel $g$ such that $g-g_*$ is PD, as \emph{general kernel}, and the general kernel $g_*$ is called \emph{indefinite} if neither of $g_*,-g_*$ is positive definite~\citep{ong2004learning}. 
General similarity and indefinite similarity are defined as well; IPDS approximates any general similarities arbitrary well. 

%According to \citet{ong2004learning}, there is a surjection between the set of such a dominated kernels $g_*$ and the set of generalized kernels defined in the vector space generated out of the cone of positive kernels.
%Considering the assumption, our idea prove Theorem~\ref{theo:universal_general} is 
% based on \citet{ong2004learning} Proposition~7 (a.k.a. generalized Mercer's theorem~\citep{chen2008generalized}). 
Our proof for Theorem~\ref{theo:universal_general} is based on Proposition~7 of \citet{ong2004learning}.
This proposition indicates that the kernel $g_*$ dominated by some PD kernels is decomposed as the difference of two PD kernels $g_+,g_-$ by considering Krein space consisting of two Hilbert spaces.
Therefore, we have $g_*(\bs f_*(\bs x),\bs f_*(\bs x')) = g_+(\bs f_*(\bs x),\bs f_*(\bs x')) - g_-(\bs f_*(\bs x),\bs f_*(\bs x'))$. 
Because of the PD-ness of $g_+$ and $g_-$, 
Theorem~\ref{theo:universal_approximate} guarantees the existence of NNs $\bs f_{\text{NN}},\bs r_{\text{NN}}$ such that 
$\langle \bs f_{\text{NN}}(\bs x),\bs f_{\text{NN}}(\bs x') \rangle$ and 
$\langle \bs r_{\text{NN}}(\bs x),\bs r_{\text{NN}}(\bs x') \rangle$, respectively, approximate 
$g_+(\bs f_*(\bs x),\bs f_*(\bs x'))$ and $g_-(\bs f_*(\bs x),\bs f_*(\bs x'))$ arbitrary well. Thus proving the theorem. 
This idea for the proof is also interpreted as a generalized Mercer's theorem for Krein space (there is a similar attempt in \citet{chen2008generalized}) by applying Mercer's theorem to the two Hilbert spaces of \citet[Proposition~7]{ong2004learning}.

\subsection{Deep Gaussian embedding}
To show another example of non-CPD kernels, Deep Gaussian embedding~\citep{bojchevski2018deep} is reviewed below.

\begin{ex}[Deep Gaussian embedding] 
\label{ex:deep_gaussian_embedding}
\normalfont 
Let $\mathcal{Y}$ be a set of distributions over a set $\bs Z \subset \mathbb{R}^q$. 
Kullback-Leibler divergence~\citep{kullback1951information} between two distributions $\bs y,\bs y' \in \mathcal{Y}$ is defined by 
$$
    d_{\text{KL}}(\bs y , \bs y')
    :=
    \int_{\bs Z} y(\bs z) \log \frac{y(\bs z)}{y'(\bs z)} \diff \bs z, 
$$
where $y(\bs z)$ is the probability density function corresponding to the distribution $\bs y \in \mathcal{Y}$.

With the same setting in Section~\ref{sec:background}, Deep Gaussian embedding~\citep{bojchevski2018deep}, which incorporates neural networks into Gaussian embedding~\citep{vilnis2014word}, learns two neural networks $\bs \mu:\mathbb{R}^p \to \mathbb{R}^q,\bs \Sigma:\mathbb{R}^p \to \mathbb{R}^{q \times q}_+$ so that 
the function $\sigma(-d_{\text{KL}}(
\mathcal{N}_q(\bs \mu(\bs x_i),\bs \Sigma(\bs x_i)),
\mathcal{N}_q(\bs \mu(\bs x_j),\bs \Sigma(\bs x_j)) ))$ 
approximates $E(w_{ij} | \bs x_i, \bs x_j)$. 
$\mathbb{R}_+^{q \times q}$ is a set of all $q \times q$ positive definite matrices and $\mathcal{N}_q(\bs \mu,\bs \Sigma)$ represents the $q$-variate normal distribution with mean $\bs \mu$ and variance-covariance matrix $\bs \Sigma$.

Unlike typical graph embedding methods, deep Gaussian embedding maps data vectors to distributions as
$$
    \mathbb{R}^p \ni \bs x \mapsto \bs y:=\mathcal{N}_q(\bs \mu(\bs x),\bs \Sigma(\bs x)) \in \mathcal{Y},
$$
where $\bs y$ is also interpreted as a vector of dimension $K=q+q(q+1)/2$ by considering the number of parameters in $\bs \mu$ and $\bs \Sigma$.
Our concern is to clarify if $d_{\text{KL}}$ is CPD. However, in the first place, $d_{\text{KL}}$ is not a kernel since it is not symmetric. 
In order to make it symmetric, Kullback-Leibler divergence may be replaced with Jeffrey's divergence~\citep{kullback1951information}
$$
d_{\text{Jeff}}(\bs y,\bs y')
:=
d_{\text{KL}}(\bs y,\bs y')
+
d_{\text{KL}}(\bs y',\bs y).
$$
\end{ex}

Although $-d_{\text{Jeff}}$ is a kernel, it is not CPD as shown in Proposition~\ref{prop:gaussian_not_cpd}. 

\begin{prop}
\label{prop:gaussian_not_cpd}
\normalfont
$-d_{\text{Jeff}}$ is not CPD on $\tilde{\mathcal{P}}_q^2$, where $\tilde{\mathcal{P}}_q$ represents the set of all $q$-variate normal distributions. 
%$-d_{\text{D-Gauss}}$ is not CPD on $\mathbb{R}^{p} \times \mathbb{R}^p$.
\end{prop}

A counterexample of CPD-ness is, $n=3,q=2,
c_1=-2/5,c_2=-3/5,c_3=1, 
\bs y_i=\mathcal{N}_2(\bs \mu_i,\bs \Sigma_i) \in \mathcal{Y} \: (i=1,2,3),
\bs \mu_1=(2,1)^{\top},
\bs \mu_2=(-1,1)^{\top},
\bs \mu_3=(1,2)^{\top},
\bs \Sigma_1=\mathrm{diag}(1/10,1),
\bs \Sigma_2=\mathrm{diag}(1/2,1),
\bs \Sigma_3=\mathrm{diag}(1,1)$. 

%Thus Deep Gaussian embedding deals with a distance which cannot be approximated by our proposed model~(\ref{eq:gij2}). 
%したがって，Deep Gaussian embeddingで表現される類似度は，CPDよりさらに広く特徴量の内積だけでは表現できないクラスに入っている．

We are yet studying the nature of deep Gaussian embedding. 
However, as Proposition~\ref{prop:gaussian_not_cpd} shows, 
negative Jeffrey's divergence used in the embedding is already proved to be non-CPD; SIPS cannot approximate it. 
IPDS model is required for approximating such non-CPD kernels. 
Thus we are currently trying to reveal to what extent IPDS applies, by classifying whether each of non-CPD kernels including negative Jeffrey's divergence satisfies the assumption on the kernel $g_*$ in Theorem~\ref{theo:universal_general}.

\end{document}